\documentclass[sigconf]{acmart}       
\usepackage{amssymb}

\usepackage[switch]{lineno}  % 解决双栏分页和行号问题

\usepackage{amsmath,amssymb,amsfonts}
\usepackage{algorithmic}
\usepackage{graphicx}
\usepackage{textcomp}
\usepackage{xcolor}

\usepackage{tabularx}
\usepackage{algorithm}
\usepackage{multirow}
\usepackage{multicol}
\usepackage{rotating}
\usepackage{algorithmic}
\usepackage{graphicx}
\usepackage{svg}
\usepackage{subfigure}
\usepackage{enumitem}
\usepackage{arydshln}
\usepackage[justification=centering]{caption}
\usepackage[compact]{titlesec}
\AtBeginDocument{%
  }

\setcopyright{acmlicensed}
\copyrightyear{2018}
\acmYear{2018}
\acmDOI{XXXXXXX.XXXXXXX}
\acmConference[Conference acronym 'XX]{Make sure to enter the correct
  conference title from your rights confirmation email}{June 03--05,
  2018}{Woodstock, NY}
\acmISBN{978-1-4503-XXXX-X/2018/06}

\begin{document}

\title{\textsf{AdaptGOT}: A Pre-trained Model for Adaptive Contextual POI
Representation Learning}

\author{Xiaobin Ren}
\affiliation{
\institution{University of Auckland}
  \city{Auckland}
  \country{New Zealand}
}
\email{xren451@aucklanduni.ac.nz}

\author{Xinyu Zhu}
\affiliation{
  \institution{University of Auckland}
  \city{Auckland}
  \country{New Zealand}
}
\email{xzhu662@aucklanduni.ac.nz}

\author{Kaiqi Zhao}
\affiliation{
  \institution{University of Auckland}
  \city{Auckland}
  \country{New Zealand}
}
\email{kaiqi.zhao@auckland.ac.nz}

\renewcommand{\shortauthors}{Trovato et al.}

\begin{abstract}
Currently, considerable strides have been achieved in Point-of-Interest (POl) embedding methodologies, driven by the emergence of novel POI tasks like recommendation and classification.  Despite the success of task-specific, end-to-end models in POI embedding, several challenges remain. These include the need for more effective multi-context sampling strategies, insufficient exploration of multiple POI contexts, limited versatility, and inadequate generalization. To address these issues, we propose the \textbf{AdaptGOT} model, which integrates both the (\textbf{Adapt})ive representation learning technique and the Geographical-Co-Occurrence-Text (\textbf{GOT}) representation with a particular emphasis on \textbf{G}eographical location, Co-\textbf{O}ccurrence and \textbf{T}extual information. The \textbf{AdaptGOT} model comprises three key components: (1) \textbf{contextual neighborhood generation}, which integrates advanced mixed sampling techniques such as KNN, density-based, importance-based, and category-aware strategies to capture complex contextual neighborhoods; (2) an advanced \textbf{GOT representation} enhanced by an attention mechanism, designed to derive high-quality, customized representations and efficiently capture complex interrelations between POIs; and (3) the \textbf{MoE-based adaptive encoder-decoder} architecture, which ensures topological consistency and enriches contextual representation by minimizing Jensen-Shannon divergence across varying contexts. Experiments on two real-world datasets and multiple POI tasks substantiate the superior performance of the proposed \textsf{AdaptGOT} model.

\end{abstract}

\received{20 February 2007}
\received[revised]{12 March 2009}
\received[accepted]{5 June 2009}
%\pagenumbering{gobble}  % 禁用第一页页码
\maketitle               
%\linenumbers   
\pagenumbering{arabic}  % 从第2页开始启用页码

\section{Introduction}
\titlespacing*{\section}{0pt}{-4pt}{4pt} % Reduces the space before this section only
%1: Highlight the necessity of POI embedding for POI recom-mendation or classification task, etc. To prove our task is valuable.—Done for now (1.19).

The recent progressions in location-based systems have enabled users to share their experiences and insights at points of interest (POIs). The accumulation and accessibility of significant footprint data serve as the foundation for advancing POI recommendations, category classification, function zone identification, and the discernment of users' living patterns~\cite{C-WARP, GeoSAN}, etc.  In this context, significant attention has been directed toward investigating POI embedding, aiming to capture intricate semantic relationships and contextual information associated with POIs to enhance diverse location-based applications and analysis. 

% With the rapid advancement of location-based systems, the convenience of collecting human mobility check-in data has significantly increased. A typical check-in encompasses various attributes, including user name, points of interest (POIs), and timestamp. Notably, POIs are characterized by several attributes, such as POI ID, geographical location, categories (e.g., gym or restaurant), and reviews\footnote{Review: Comments of users to the POIs}. Learning the comprehensive context from check-ins can yield benefits not only for various downstream applications, such as POI recommendations, category classification, function zone identification, and users' living pattern recognition\cite{C-WARP}\cite{LBSN2Vec}\cite{GeoSAN}, etc., but also for transferable tasks, given the sparse distribution of check-ins in many cities\cite{ST-TransRec}\cite{PACE}\cite{PR-UIDT}.
 POI embedding plays a pivotal role in various downstream tasks. A wide range of contextual information has been considered in POI embedding, such as geographical coordinates, POI sequences, spatial co-occurrence patterns of POIs, textual information, and additional auxiliary features. 
POI2Vec \cite{POI2Vec} and Hier-CEM \cite{Hier-CEM} leverage geographical coordinates to embed spatially organised POI pairs, modelling complex spatial correlations for category prediction tasks. Similarly, HMRM \cite{HMRM} and GeoSAN \cite{GeoSAN} utilise user trajectory data to capture POI co-occurrences. Besides, recent work \cite{zhang2020accurate} incorporates contextual information through an additional content layer to integrate textual data.

\begin{figure}
\vspace{-1em}
    \centering
    \includegraphics[width=\linewidth]{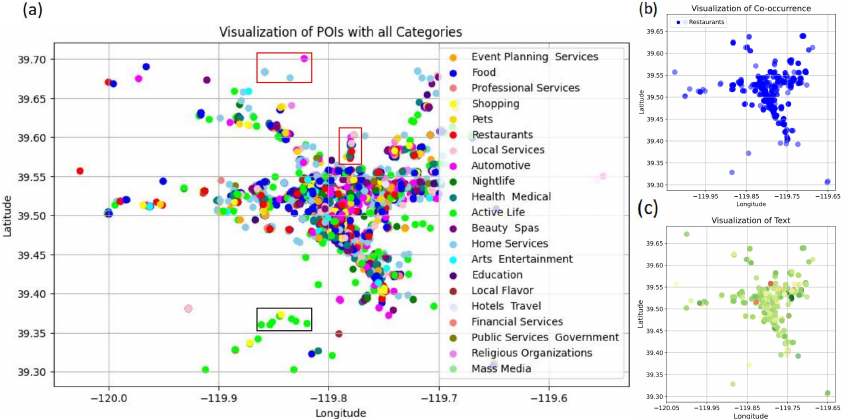}
        \caption{(a) Visualization of POIs on Yelp Nevada dataset. (b) Co-occurrence of all POIs, darker blue indicates frequent co-selection with other POIs. (c) Sentiment analysis of all POIs, with orange representing negative evaluations.}
    \label{fig:Motivation_of_GOT_and_subgraphs}
    \vspace{-2em}
\end{figure}

Despite the notable achievements of contemporary methodologies, several limitations warrant further investigation. 

\textbf{First, current research lacks consideration of diverse sampling strategies under multiple POI contexts (Gap~1).} Existing POI-based graph models fall into three categories: Expressive Graph Neural Networks (GNNs)\cite{HighorderGNN,HighorderGemb}, Message Passing Neural Networks (MPNN)\cite{MSPGNN,MSGGCN}, and subgraph-based approaches~\cite{NAVIP}. 
% MPNNs are constrained by the first-order Weisfeiler-Lehman (1-WL) test, limiting their ability to capture essential structural information~\cite{PowerfulGNNs}.
While more expressive k-WL GNNs~\cite{HighorderGNN} address this limitation, their reliance on $O(k)$-order tensors restricts scalability for large graphs. Recent studies have explored subgraph sampling to balance expressive power and scalability~\cite{Subrecommender,SuGeR}, yet the integration of multiple contexts remains limited, introducing randomness in sampled graphs~\cite{NAVIP,ladies,bns}. Importantly, leveraging multiple subgraphs has been shown to match the expressive power of MPNNs when using sufficient layers and injective functions~\cite{PowerfulGNNs,Stars2subgraphs}. 
% Despite the advancements, current POI representation learning frameworks have yet to fully leverage the potential of multiple subgraphs to enhance representation capability, leaving a significant gap in exploiting multi-subgraph approaches under multi-context scenarios~\cite{CAPE,Meta-SKR}. Neglecting multiple contexts risks over-smoothing in the embedding space~\cite{Huangchao_GAug}, which can undermine representation quality. Incorporating diverse contexts not only mitigates this issue but also contributes to a more robust model design.
Therefore, current POI representation frameworks fail to fully exploit multi-subgraph approaches under multi-context scenarios~\cite{CAPE,Meta-SKR}, risking over-smoothing in the embedding space~\cite{Huangchao_GAug}. Incorporating diverse contexts mitigates this issue and enhances model robustness.

\textbf{Second, there is a lack of comprehensive investigations of multiple POI contexts simultaneously (Gap~2).} Existing research predominantly focuses on specific contexts, such as geographical locations, categories, co-occurrence, or textual information. However, none of the current methods fully integrate these POI contexts, limiting their broader applicability. As illustrated in Figure~\ref{fig:Motivation_of_GOT_and_subgraphs}(a), geographical location demonstrates diverse spatial correlations with categories, with certain areas exhibiting a higher density of various categories. Additionally, Figures~\ref{fig:Motivation_of_GOT_and_subgraphs}(b) and (c) highlight the distinct correlations of co-occurrence and textual information with geographical location. These intricate interrelations necessitate a comprehensive and holistic investigation of multiple POI contexts simultaneously.
%where the spatial distribution of POI categories reveled the geographical location shows a strong correlation with POIs. However, Figure~\ref{fig:Motivation_of_GOT_and_subgraphs} (b) and (c) highlight the controversy surrounding textual sentiment analysis of reviews and the co-occurrence of POIs. These inconsistencies further emphasize the necessity for a thorough investigation and nuanced understanding of the intricate patterns inherent in diverse POI contexts.
%

\textbf{Third, fixed pre-trained POI embeddings encounter significant limitations when applied to diverse downstream tasks (Gap~3).} For instance, the task of POI categorization requires an analysis of spatially proximate POIs, whereas the next POI recommendation might prioritize POIs that are visited within a short time frame~\cite{contextpoi}. However, the fixed pre-trained POI embeddings generated by existing methods lack the flexibility to adapt effectively to tasks that demand distinct contextual information.
%In this case, there is an urgent need for a highly adaptable and unified approach that can automatically and flexibly accommodate diverse contextual neighborhoods.
%
% Third, \textbf{existing methods lack generalizability to new cities or regions.} 
% Unlike pre-trained language models, which employ a fixed vocabulary, pre-trained POI embeddings often struggle when applied to new cities or regions with entirely different POIs. While transfer learning techniques~\cite{ST-TransRec,PACE,PR-UIDT} have been explored to adapt the pre-trained models to new cities, there has been limited focus on pre-training POI embedding models in an inductive manner that would allow them to be applied to new cities without requiring fine-tuning.

To address the aforementioned challenges, we introduce \textsf{AdaptGOT}, which builds upon a pre-trained \textsf{adapt}ive representation model through Mixture-of-Expert (MoE) with the integration of Geographical-Co-Occurrence-Text (\textbf{GOT}) representations. 
To address \textbf{Gap~1}, we propose a \textbf{mixed sampling subgraphs} considering multiple contexts and diverse POI neighborhoods. This design can naturally encode the locally induced subgraphs and will uplift the base model of modified GATs. The mixed sampling scheme based on several subgraphs will generate new node embeddings to overcome the over-smoothing (Section~\ref{sec:Case}) and boost the expressive power theoretically. 
To address \textbf{Gap~2}, we propose a \textbf{GOT representation module} to acquire 
three predominant representations: co-occurrence, geographical, and texts. These representations are further transformed in the \textbf{GOT attention module}, which generates the latent node and neighborhood representations in an inductive manner. 
To address \textbf{Gap~3}, we propose a specialised \textbf{adaptive encoder-decoder framework} featuring an MoE-based mechanism that dynamically adjusts POI representations across tasks without retraining. This is achieved by minimising the Jensen-Shannon (JS) divergence between sampled subgraph and raw graph distributions. The \textbf{adaptive representation aggregator} maintains subgraph topological integrity while enhancing contextual expressiveness, thereby facilitating multi-contexts decoding across POI tasks.

We summarize the contributions of this research as follows:

\begin{itemize}[leftmargin=*, itemsep=0pt, topsep=0pt]

\item Mixed Subgraph Sampling Scheme: The mixed sampling strategy integrates multiple contexts and diverse POI neighborhoods. This approach encodes locally induced subgraphs and enhances the expressive power of the base model (modified GATs) by generating new node embeddings from multiple subgraphs.

\item GOT Representation and GOT Attention Module: The GOT representation module captures three key aspects of POI data: co-occurrence, geographical, and texts. These representations are further processed using a GOT attention module to generate latent node and neighborhood embeddings in an inductive manner.

\item Adaptive Encoder-Decoder Architecture: This architecture effectively selects diverse contextual neighborhoods while preserving the topological consistency of subgraphs through the minimization of Jensen-Shannon (JS) divergence, enhancing the contextual and structural representation capabilities.

% \item Experiments: Extensive experiments on the Yelp and Foursquare datasets demonstrate the superior performance of \textsf{MOEGOT} compared to other methods. 
\end{itemize}

\vspace{8pt} % Slightly reduces the space, adjust as needed
\section{Related Work}

\subsection{POI Contextual  Learning}
% Significant efforts in recommender systems have focused on enhancing POI embeddings~\cite{C-WARP,CatDM,CEMXEM}. Recent research has focused on integrating diverse contextual factors into POI embeddings. GeoSAN~\cite{GeoSAN} employs a hierarchical grid and a self-attention-based encoder to capture geographical information. CAPE~\cite{CAPE} uses a check-in text context layer to capture textual information and POI relationships. HMRM~\cite{HMRM} models POI and POI-time co-occurrences within user trajectories, leveraging trajectory attributes. Other studies explore broader spatio-temporal impacts on POI embeddings~\cite{Getnext,Meta-SKR,NeuNext}.

Recent advancements in recommender systems have focused on enhancing POI embeddings by integrating diverse contextual factors, including geographical, textual, and spatio-temporal information~\cite{GeoSAN, CAPE, HMRM}. Approaches such as hierarchical encoders, text-based context layers, and trajectory modeling highlight the importance of multi-contextual integration in improving POI representations~\cite{C-WARP, Getnext, Meta-SKR}.

% C-WARP~\cite{C-WARP} captures temporal contexts using a skip-gram model, CatDM~\cite{CatDM} employs LSTM-based encoders for users' temporal preferences, and CEM/XEM~\cite{CEMXEM} use category-aware embeddings to identify temporal patterns in check-in data.
%The availability of these POI embedding techniques allows systems to understand and leverage the underlying patterns and structures in POI data more effectively.

%Additionally, noteworthy research efforts have been devoted to investigating the wider spatio-temporal impact on POI embeddings, as highlighted in studies such as Getnext \cite{Getnext}, Meta-SKR \cite{Meta-SKR}, and NeuNext \cite{NeuNext}. In fact, the integration of abundant contextual information has led to improvements in the model's accuracy.

To improve the generalizability of POI embeddings, new studies have explored pre-training techniques. For example, MGeo \cite{mgeo} has pioneered a pre-trained POI embedding technique that integrates geographic context to extract multi-modal correlations crucial for precise query-POI matching. 
CTLE~\cite{CTLE} devises a pre-trained method that computes a location's representation vector by considering its specific contextual neighbors within trajectories. 
SpaBERT \cite{SpaBERT}  offers a versatile geo-entity representation based on neighboring entities in geospatial datasets. However,  CTLE bias towards the next POI recommendation task. While SpaBERT demonstrates generalizability to cross-city tasks, it produces a fixed POI embedding, which inevitably impacts the model's accuracy and effectiveness in diverse tasks emphasizing distinct contexts. More recently, M3PT was proposed as a contextual POI learning framework. While its fusion of textual and visual features achieves promising results, its heavy reliance on image modality and lack of sequential modelling limit its generalisability and practical utility~\cite{m3pt}.
%For example, GETNEXT \cite{Getnext} constructs a trajectory flow map with a Graph Transformer structure to incorporate the same fragment in different trajectories within POI embeddings. Moreover, in order to integrate sequential, spatio-temporal, and social knowledge into POI embeddings, Meta-SKR \cite{Meta-SKR}, a meta-learned sequential-knowledge-aware recommender, is established through an end-to-end attention-based knowledge graph meta-learning framework. 

% More recently, the pre-training POI embedding has gained popularity in location-based system development given the great stride achieved in natural language processing (NLP) domain. Despite the great attempt in 

% However, the intrinsic reliance of user-specific attribute restrict MGEO to across city tasks\cite{Spabert}\cite{MGEO}.

\subsection{Sampling models}

In the realm of sampling, it has been well acknowledged that the application of Graph Neural Networks (GNNs) to large-scale graphs presents considerable challenges due to the substantial computational and memory costs associated with the training process \cite{bns}. As a result, numerous researchers have undertaken efforts to develop simplified models aimed at expediting the training process. These models can be broadly categorized into three main types: node-wise sampling \cite{MVSGNN}\cite{stochasticgcn}\cite{bns}, layer-wise sampling \cite{fastgcn}\cite{ladies}\cite{adaptiveGCN}, and subgraph sampling \cite{clustergcn}\cite{graphsaint}. In node-wise sampling, a set of neighbors is sampled for each node. Layer-wise sampling involves selecting layer-dependent neighbors based on the calculation of importance for different nodes. While each method has demonstrated efficiency in sampling, they lack the ability to integrate different graph topologies effectively. Furthermore, in the context of POI, existing research fails to determine appropriate graph topologies when considering multiple POI contexts \cite{Getnext}.

\vspace{8pt} 
\section{Problem Definition \& Theoretical Analysis}

%We developed \textsf{MOEGOT}, a pre-trained POI embedding model designed to enable inductive learning of unified and general POI representations, capable of accommodating various tasks across diverse contexts. \textsf{MOEGOT} incorporates several novel components: 1) a mixed sampling technique that generates distinct contextual neighborhoods for POIs; 2) a GOT representation layer and a GOT attention layer that extract and aggregate essential information from multiple contexts; and 3) an MOE-based module that effectively selects and combines contextual embeddings from different sources. Figure~\ref{fig:Architecture_MOEGOT} illustrates the overall architecture of \textsf{MOEGOT}.

% Next, we proceed to define the POI representation learning problem and elaborate on the essential components of \textsf{MOEGOT}.
\begin{figure*}[htbp]
    \centering
    \includegraphics[width=\linewidth]{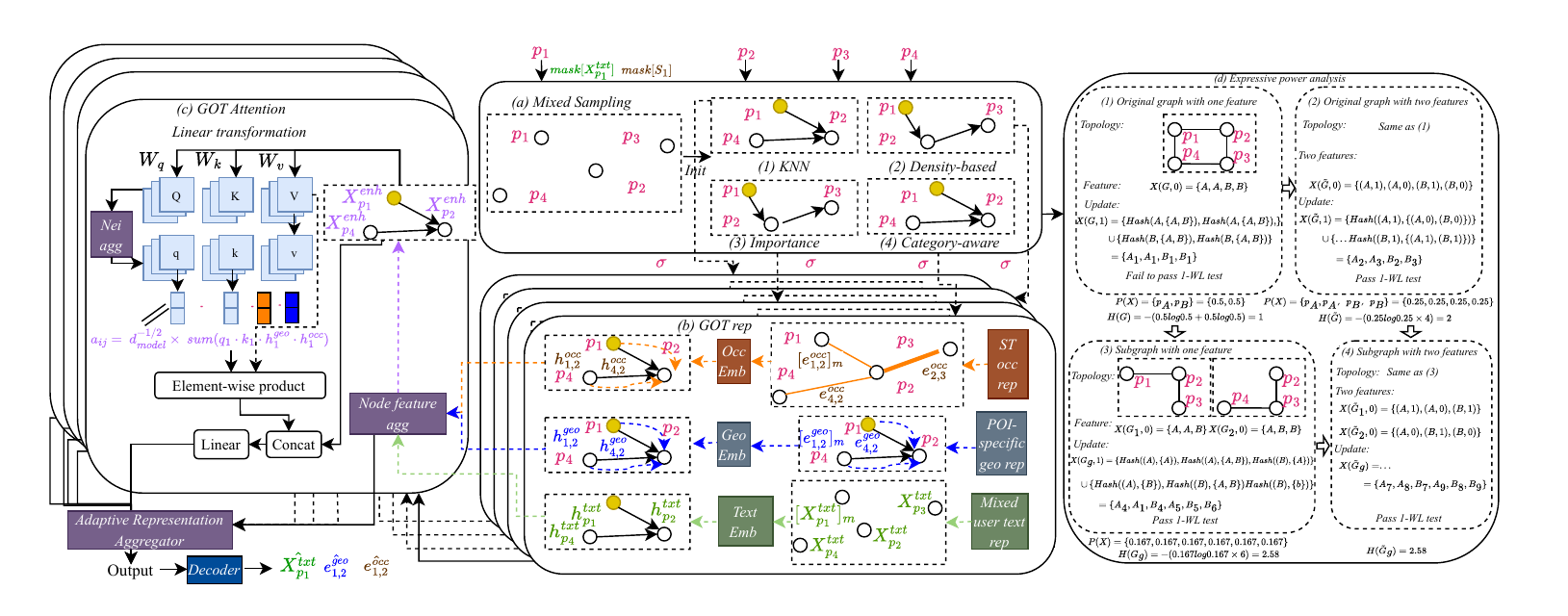}  % Adjusted width to minimize the figure
    \caption{Architecture (a), (b), (c) and expressive power (d) of \textsf{AdaptGOT}. In (d), we initialize one or two features across different graph topologies to demonstrate the 1-WL test and examine the entropy under each scenario on Section~\ref{subsec:Expre}.}
    \label{fig:Architecture_MOEGOT}
    \vspace{-1em}  % Reduce space below the figure (optional).
\end{figure*}

\subsection{Problem definition}
Let $P=\{p_{1},p_{2},p_{3},...,p_{N}\}$ be a set of POIs, $U=\{u_{1},u_{2},u_{3},...,u_{M}\}$ denote a set of users and $T=\{t_{1},t_{2},t_{3},...,t_{K}\}$
signify a set of timesteps, where $N,M,K$ denote the positive integers. Each POI $p \in P$ is characterized by a tuple  $p=<lat, lon, cat>$ comprising latitude, longitude, and category. A check-in is defined as a tuple $q=<u,p,t> \in U\times P \times T$, indicating that users $u$ visit $p$ at timestep $t$. We denote the set of all check-ins by $Q$.
%For the $i$-th user, a check-in can be represented by $s_{i}=<p_{i},t_{i}> \in P \times T$, and the complete set of historical check-ins for the $i$-th user over all historical timesteps $T$ is represented by $S_{i}=<(p_{1},t_{1}),(p_{2},t_{2}),...,(p_{n},T)>$. 
% Let $G = (\mathcal{V}, \mathcal{E})$ be a graph, each node $\mathcal{V}$ represents a POI. Each edge in $\mathcal{E}$ represents a connection between two POIs, and it is learnable~\cite{ENADPool}. 
%$\mathcal{N}_{(p_{i})}$ stands for the neighborhoods for $p_{i}$. We use geographical embedding and co-occurrence embedding as edges. 
Let $X_{u_{i},p_{j}}^{txt}$ denote the review texts from user~$i$ pertaining to POI $p_{j}$, where $i$ represents the $i$-th user and~$j$ represents the $j$-th POI. The objective of \textsf{AdaptGOT} is to generate pre-trained embeddings $H=f(X_{U,P}^{txt},P, Q)$ for all POIs following the self-supervised manner. 

\subsection{Expressive power analysis}
\label{subsec:Expre}

Existing studies have proved the effectiveness of the graph formulation of POI data in learning POI representations~\cite{Huangchao_GAug}\cite{GraphPOIAAAI}. Following prior work~\cite{ENADPool}, we construct a POI graph $G=(\mathcal{V}, \mathcal{E})$ where each node in $\mathcal{V}$ is a POI, and the edges in $\mathcal{E}$ are learnable and determined by the node features. However, as discussed in Section 1, existing MPNNs and subgraph-based models may not be adequate for learning robust and effective POI representations. To obtain a theoretical view of the problem, we will first analyze the bottleneck at current MPNNs and subgraph-based methods. Subsequently, we will theoretically justify the need for mixed sampling strategies to improve the expressiveness of the learned embedding. 

\subsubsection{MPNNs}

Let $h_{p_{i}}^{l}$ denotes the embedding of $p_{i}$, then the MPNNs can be described as follows: 

\begin{align}
h_{p_{i}}^{(l+1)} &= \phi^{(l)}\left(h_{p_{i}}^{(l)}, f^{(l)}\left({h_{p_{j}}^{(l)} \mid u \in N(v)}\right)\right); \, l=0,\ldots,L-1; \label{eq:MPNN1}\\
h_{G} &= \text{POOL}\left({H_{v}^{l} \mid v \in \mathcal{V}}\right),
\label{eq:MPNN2}
\end{align}
where $h^{(0)}_{p_{i}}=x_{i}$ is the constructed feature vector of POI $p_{i}$, $\phi$ represents the injective function.
The expressive power of MPNNs' upper bound from its close relation to the 1-WL isomorphism test~\cite{HighorderGemb}. Note that the current MPNNs rely heavily on the star graph with limited features, which makes the star not distinguishing enough~\cite{Stars2subgraphs}. In view of that, the design of the subgraph under multiple contexts will facilitate the isomorphism test.

\subsubsection{1-WL isomorphism test}\label{sec:WL_test}

The 1-WL is a widely used framework in graph theory to determine the equivalence of two graphs by iteratively updating node embeddings based on local neighborhood information. 
The node embeddings at iteration 
$t+1$ are updated as: 

\begin{equation}
h_v^{(t+1)} = \text{Hash}\left( h_v^{(t)}, \text{multiset} \{ h_u^{(t)} : u \in \mathcal{N}(v) \} \right)    
\end{equation}

The process begins with initial attributes $h_{v}^{(0)}$, which typically represents node attributes or graph-invariant properties such as degree. Hash$()$ denotes the Hash function.
\subsubsection{Multiple contexts}\label{sec:Multi_contexts}
Next, we demonstrate that using multiple contexts enhances expressive power through Theorems 1 and 2. Theorem 1 shows that multiple contexts reduce feature duplication, improving the 1-WL test's discriminative power. Theorem 2 proves that integrating multiple contexts increases node embedding entropy. Figure~\ref{fig:Architecture_MOEGOT} (d) illustrates these findings by an example.

\textbf{Theorem 1}\footnote{Detailed proof on Theorem 1, 2, and 3: \url{https://anonymous.4open.science/r/AdaptGOT-2748/README.md}}: \emph{Incorporating multiple contexts reduces the probability of feature duplication among nodes, thereby enhancing the discriminative power of the 1-WL test}. Specifically, 
\begin{equation}
P_{\text{conflict}}^{\text{single}} \geq P_{\text{conflict}}^{\text{multi}},    
\end{equation}
where \( P_{\text{conflict}}^{\text{single}} \) and \( P_{\text{conflict}}^{\text{multi}} \) represent the probabilities of feature conflict between two distinct nodes in single-feature graphs and multi-feature graphs, respectively. 

We provide the proof by analyzing the feature space size in multiple-feature case grows exponentially with dimension size $k$, significantly reducing the likelihood of conflicts. The (1) and (2) of Figure~\ref{fig:Architecture_MOEGOT} (d) also support our claim. Utilizing additional feature will generate discriminative label $X(\tilde{G},1)$ through the first update. However, using one feature will bring duplicated labels.

\textbf{Theorem 2}. 
\emph{Incorporating POI contextual information into node representations increases the entropy of the node embedding.}

\begin{equation}
H(\tilde{G})\geq H({G}),
\end{equation}
where $H(\tilde{G})$   represents the entropy under multiple features.

% \textbf{Proof:} The entropy of a probability distribution \( p(x) \) is given by:
% \[
% H(p) = -\sum_{x} p(x) \log p(x).
% \]
% For the label distributions \( p_\text{struct}(x) \) (without features) and \( p_\text{features}(x) \) (with features), we have:
% \[
% H(h_v^{(0)}) = 
% \begin{cases} 
% -\sum_{x} p_\text{struct}(x) \log p_\text{struct}(x), & \text{without features,} \\
% -\sum_{x} p_\text{features}(x) \log p_\text{features}(x), & \text{with features.}
% \end{cases}
% \]

% Since \( p_\text{features}(x) \) subsumes \( p_\text{struct}(x) \) by incorporating feature variability:
% \[
% p_\text{features}(x) = \sum_{y} p_\text{struct}(x \mid y) p_\text{features}(y),
% \]
% the additional variability implies:
% \[
% H(h_v^{(0)}_\text{with features}) \geq H(h_v^{(0)}_\text{without features}).
% \]

% Thus, incorporating contextual information enriches node embeddings and increases entropy, leading to more informative initial representations.

% \subsubsection{Multiple contexts}\label{sec:Multi_contexts}

% We propose to leverage multiple contexts of POIs, such as geographical contexts and co-occurrence contexts. Next, we will then introduce why enriching node features enhances subgraph expressive power through 1-WL isomorphism test.

We prove Theorem 2 by comparing entropy before and after incorporating additional features. As shown in Figure~\ref{fig:Architecture_MOEGOT} (d), the entropy increases from \( H(G) = 1 \) to \( H(\tilde{G}) = 2 \).

\subsubsection{Subgraphs under multiple contexts}

To mitigate the problems of existing subgraph-based methods, we propose to leverage multiple contexts of POIs. As a result, the node features can be enriched by these contexts. Next, we will then introduce why enriching node features enhances expressive power by 1-WL isomorphism test.

\textbf{Theorem 3.}
\emph{Leveraging multiple subgraphs enhances entropy and improves label discrimination.}

The entropy of the node label distribution after the \( t \)-th update in the 1-WL test satisfies:
\begin{equation}
P_{\text{multi}}^{(t)} \geq P_{\text{single}}^{(t)}, 
H_{\text{multi}}^{(t)} \geq H_{\text{single}}^{(t)}. 
\end{equation}
where
$P_{\text{single}}^{(t)}$ and $H_{\text{single}}^{(t)}$ stands for the entropy and probability of the label conflict of a single graph, 
while $P_{\text{multi}}^{(t)}$  and $H_{\text{multi}}^{(t)}$ represent the entropy and label conflict of multiple subgraphs. The proof follows the similar process of Theorem 1 and 2. Also, we conclude that the entropy $H(\hat{G}_{g})=2.58$ in Figure~\ref{fig:Architecture_MOEGOT} (d) for subgraphs ${G}_{g}$ under single feature and multiple features, which shows that applying multiple subgraphs will bring better representative power. 
% Therefore, we believe that the consideration of multiple POI contexts under multiple subgraphs is technically sound from the expressive power analysis. We will then provide the methodology for the details of the architecture.
\vspace{8pt} % Slightly reduces the space, adjust as needed
\section{Methodology}
In this work, we propose \textsf{AdaptGOT}, a unified framework for POI representation learning that addresses three key challenges: the need for scalable context-aware sampling, unified multi-context POI representations, and adaptive subgraph selection for POI tasks. \textsf{AdaptGOT} comprises three core components. First, a mixed subgraph sampling strategy integrates diverse POI contexts to enhance expressive power. Second, the GOT representation and attention module unifies three POI contexts, while its attention mechanism inductively generates latent node and neighborhood embeddings. Finally, the architecture adaptively selects contextual neighborhoods and preserves topological consistency through JS divergence minimization, improving structural and contextual representations. Figure~\ref{fig:Architecture_MOEGOT} shows the architecture of \textsf{AdaptGOT}.

\subsection{Contextual Neighborhood Identification via Mixed Sampling}
% \textbf{1.Introduce Why do we need Initialization 2.Briefly Introduce current methods' solution of initialization. Don't need to be very in detail. 3. Very briefly Introduce the same/difference between ours and theirs(Like GETNEXT, we have some difference, introduce the difference and why cause this difference) 4. Introduce the process of our init, equations should be attached as well.}
% dataset setup move to here, briefly discuss different dataset that has review information or not. can initial the dataset with review, also can inital teh dataset without review info
To effectively capture diverse POI contextual information, we propose a comprehensive sampling scheme incorporating four strategies. KNN and density-based sampling address geographical proximity, with KNN focusing on local neighbors and density-aware sampling capturing broader spatial correlations. Importance sampling leverages GOT contexts to prioritize significant POI interactions, while category-aware sampling considers categorical similarities. Notably, this is the first work to introduce multiple sampling schemes for POI embeddings considering multiple contexts.

% \kaiqi{What do you mean by contexts provided by ``the GOT'', what is ``the GOT''??}\XB{we explain GOT in the introduction part} 

% Differently, KNN constructs from local neighbors perspective but density-aware constructs the neighbors from a global view. As for importance sampling and category-aware sampling, importance sampling uses both textual information, co-occurrence information and geographical information. It calculates high score of neighbors through combining three different contexts. Category-aware sampling determines the important neighbors from calculating the highest score under  

% We initialize the Geographical information using the latitude, longitude of each POI, and also integrate the check-in sequence from each user. In order to capture the ,we draw inspiration from GETNEXT when processing the check-in sequence which considers check-in sequences within fixed time frames (e.g., every 24 hours), essentially segmenting user activities into short chunks.
% We diverge from this by segmenting check-in sequences based on a threshold duration of inactivity. If the interval between two check-ins exceeds 24 hours, we treat the subsequent check-in as the start of a new sequence.

% Therefore, we construct the $S_{i}^{sep}=\{<(p_{1},t_{1}),\\
% (p_{2},t_{2}),...,(p_{n},t_{sep_{1}})>;<(p_{sep_{2}},t_{sep_{2}}),\\(p_{sep_{3}},t_{sep_{3}}),...,(p_{T},t_{T})>\}$, where $sep$ represents the number of separation in the continuous time frames. 

\subsubsection{KNN sampling}
The KNN sampling method selects a fixed number \( k \) of the nearest neighbors \( \mathcal{N}_{knn}(p_{i}) \) for a given POI \( p_{i} \), based on geographical proximity. Here, \( \mathcal{N}_{knn} \) denotes the set of neighboring POIs identified by the KNN algorithm.

  % Adjust the space above the figure (negative value reduces space)
% \begin{figure}[h!]
%     \centering
% \includegraphics[height=2cm]{Fig/Example_knn_density.png}
%     \caption{Example of KNN and density sampling. Three polygons represent the three users' historical check-in records for target point (red dot).}
% \label{fig:Example_knn_density}
% \end{figure}

\subsubsection{Density-based sampling}
To account for the visiting preferences of different users towards specific $p_{i}$, we calculate the density across all historical moments \( T \) and then identify the top-\( k \) neighbors. Formally, this can be expressed as:
\begin{equation}
   \mathcal{N}_{den}(p_{i}) = \text{Topk}(Den(p_{i},g(p_{i}))),
\end{equation}
where \( \mathcal{N}_{den} \) denotes the density-based neighboring set, $g$ function returns the correlated historical check-in of $p_{i}$, and \( Den \) represents the density function, which is defined as:
%which returns the density score for all POIs in three polygons.
%The \( Den \) can be formulated as follows:

\begin{equation}
\hat{Den}(\mathbf{x}) = \frac{1}{|\mathcal{N}_{den}| b^2} \sum_{n=1}^{|\mathcal{N}_{den}|} K\left(\frac{p^{\prime} - p^{\prime}_{n}}{b}\right),
\label{eq:Den}
\end{equation}
where $p^{\prime}_{n}$ represents the  subset of $p$ that includes only the latitude and longitude on the $n$-th POI, $b$ denotes the bandwidth and $K$ denotes the the non-parametric Gaussian kernel\cite{KernelSmoothing}.

\subsubsection{Importance sampling}
The importance sampling has been studied by many works~\cite {ladies,fastgcn} while complicated POI contexts are neglected, like positive remarks of reviews.
Our importance sampling method aggregates textual information \( X_{p_{i}}^{txt} \), the frequency matrix \( X_{p_{i}}^{freq} \), and the distance matrix \( D_{u_{i}} \). It is defined as:
\begin{equation}
\mathcal{N}_{imp}(p_{i})=Topk\left(-\frac{X^{freq}_{p_{i}} \odot Senti(X^{txt}_{*;p_{i}}) + \gamma}{D_{u_{i}}+\gamma}\right),
\label{equ:imp_sampling}
\end{equation}
where \( X_{p_{i}}^{freq} \) represents the frequency matrix of all users in $p_{i}$, $X^{txt}_{*;p_{i}}$ represents the set of reviews from all users regarding POI $p_i$, and $Senti$ denotes the sentiment analysis function, which returns a positive, negative, or neutral evaluation of a review by analyzing the frequency of various words~\cite{textblob}, as illustrated in Figure \ref{fig:Motivation_of_GOT_and_subgraphs}(c). The symbol \( \odot \) indicates the element-wise product and a small value \( \gamma \) is added to prevent zero denominators.

\subsubsection{Category-aware sampling}
Category-aware sampling strategy bears resemblance to the approach delineated in Equation (\ref{equ:imp_sampling}). Formally, this strategy is formulated as follows:
\begin{equation}
   \mathcal{N}_{cat}(p_{i})=Topk(-\frac{X^{freq}_{u_{i}} \odot X^{cat}_{u_{i}}+\gamma}{D_{u_{i}}+\gamma}), 
   \label{equ:cat_sampling}
\end{equation}
where $X^{cat}_{u_{i}}$ denotes the category-based matrix, which computes the normalized frequency matrix of the category.

% \noindent 4) \textbf{Contextual graph}. 
% We construct the contextual directed graph based on four sampling strategies\footnote{We use directed graph because of the different directions in co-occurrence representation}, we use $p_{i}$ represents the node for simplicity, the edge is defined by $e$.  
\subsection{GOT representation}
\label{sec:GOT representation}
% \textbf{1.Introduce the connection between init and GOT representation. Like: After Init, we got..., which is highly needed in GOT rep...} \textbf{2.Introduce the motivation again of GOT representation, should be consistent with introduction.3.Introduce our methods in detail.Remember to introduce each component within our figure.
% }

After capturing comprehensive contextual neighborhoods using the mixed sampling scheme, the GOT representation is proposed to create a unified representation that fully captures the intricate interrelationships among location, co-occurrence, and category information simultaneously.

\subsubsection{Spatio-temporal Co-occurrence representation} 

% Contemporary POI recommendation models rely extensively on historical user check-in data, frequently resorting to direct time information or time slots to depict check-in hours \cite{LSTPM}. Conversely, our objective is to leverage historical user check-in data comprehensively for cross-city tasks while eliminating the limitations imposed by specific time encoding methodologies. To achieve this, we devise a directed occurrence encoder to acquire representations from the check-in sequence, specifically tailored for cross-city tasks.

We selected co-occurrence as one of the POI contexts because it can be combined with historical POIs to measure general user preferences over a long time span. Additionally, it eliminates the need for time encoding in models like transformers, making the model suitable for zero-shot cross-city scenarios (Experiment on Section~\ref{sec:othertask}).

Giving $S_{i}$ as the complete set of historical check-ins for $i$-th users over timesteps $T$, the co-occurrence representation can be formulated as :
\begin{equation}
    O_{i,j} = \frac{\#Occ_{i,j}}{\#S_{i}},
\end{equation}
where $\#Occ_{i,j}$ stands for the number of co-occurrence between $p_{i}$ and $p_{j}$, $S_{i}$ serve to normalize the co-occurrence $Occ_{i,j}$ to obtain normalized co-occurrence $O_{i,j}$. 
% Note that $O_{i,j}$ is not the same as $O_{j,i}$ in the directed graph representation.

% It's important to note that a higher count for a location appearing earlier in sequences not only indicates its sequence precedence but also implies its relative popularity or the frequency of check-ins by users. Since our focus is solely on capturing the historical check-in relationships between locations, we do not account for locations that follow others in the sequence. Therefore, \( O_{i,j} \) is distinct from \( O_{j,i} \), as \( O_{i,j} \) captures the information when \( i \) was in the historical check-in sequence of \( j \). We incorporate \( O_{i,j} \) into our directed graph representation accordingly. 
% Consider a user's check-in sequence denoted as \( l_1, l_2, l_3 \). Our goal is to capture the sequence dynamics where a location \( l_i \) is visited before \( l_j \), reflecting scenarios like visiting a cafe before a movie theater. To encode this sequence information, we create tuples \( [l_i, l_j] \) whenever \( l_i \) precedes \( l_j \) in the sequence. Subsequently, we tally the frequency of each tuple across all user sequences, which is then normalized by dividing the count by the total number of sequences to yield a relational representation \( O_{i,j} \) between locations, as expressed in the equation:
% \begin{equation}
%     O_{i,j} = \frac{\#Occ_{i,j}}{\#\text{Total Sequences}}
% \end{equation}

\subsubsection{POI-specific Geographical representation and embedding}
% To capture the inductive geographical representations between two stations, existing methods establish an absolute position embedding (APE) as the token embedding for the model input. However, due to the strong influence of current cities's geographic information on APE, transferable embedding techniques are not applicable. Furthermore, simple addition can lead to noise aggregation and mixed information issues. In light of this, we opt for a relative location geographical representation approach. 
%This method represents relative distance and azimuth through \textbf{POI-specific geo representations} and then transforms them into a hidden state using \textbf{geographical embeddings (GeoEmb)}, as illustrated by the blue square in figure~\ref{fig:Architecture_MOEGOT}.

%\noindent\textbf{POI-specific geo representation}.
Unlike SpaBERT\cite{SpaBERT}, we design a POI-specific geographical representation that integrates a relative position encoding. 
% As portrayed in Figure ~\ref{fig:Geo encoding}, SpaBERT\cite{SpaBERT} computes relative distances between adjacent POIs, which can result in remote stations having identical embeddings, as demonstrated by $d_{p1,p4}$ and $d_{p1,p3}$ in the left part of Figure ~\ref{fig:Geo encoding}. In contrast, our approach integrates both azimuth and distance to generate a unique geographical encoding for each $p_{i}$. 
%
Specifically, given a pair of POIs $p_{i}, p_{j}$, the relative position of two POIs is captured using the following equations:
$\boldsymbol{r}_{ij}=[s,\theta_{1}],$
$\boldsymbol{r}_{ji}=[s,\theta_{2}],$
where \( s \) represents the distance, and \( \theta_{1} \) and \( \theta_{2} \) denote the azimuth from \( p_{i} \) to \( p_{j} \) and from \( p_{j} \) to \( p_{i} \) respectively. 
% \vspace{-0.5cm}  % Adjust the space above the figure (negative value reduces space)
% \begin{figure}[h!]
%     \centering
% \includegraphics[width=0.55\linewidth]{Fig/Diff geo encoding.pdf}
%     \caption{Different geographical encoding strategy}
%     \label{fig:Geo encoding}
% \end{figure}

With the relative position vectors $\boldsymbol{r}_{ij}$, we employ a two-layer FCN to obtain the geographical representation of a POI $p_i$ given its geographical neighbor $p_j$:
\begin{equation}
    c_{ij}=(\boldsymbol{r}_{ij}W_{r}^{1}+b_{r}^{1})W_{r}^{2}+b_{r}^{2},c_{ij} \in R^{d_{model}},
\end{equation}
where $W^1_{r}, W^2_{r}, b^1_{r}$ and $b^2_{r}$ are learnable parameters.

\subsubsection{Mixed user text representation}
We formulate a composite user text representation alongside a BERT text embedding~\cite{BERT} to effectively capture textual information. Since each POI receives a range of reviews, both positive and negative, from different users, we aggregate all users' reviews for each POI separately to form a unified evaluation of the POI. Formally, \( X_{p_{i},u_{j}}^{txt} \) represents the review of user \( j \) towards the \( i \)-th POI.
\begin{equation}
\label{eq:fusetext}
X_{p_{i}}^{txt}=Concat\{X_{p_{i},u_{1}}^{txt},X_{p_{i},u_{2}}^{txt},...,X_{p_{i},u_{M}}^{txt}\}   
\end{equation}
where \( X_{p_{i}}^{txt}\) denotes the representation of each POI across all users with regard to integrated textual information.

\subsubsection{Text embedding}
We adopt a $BERT_{base}$ as the encoder of text embedding to reflect the sophisticated sentiment, preference and descriptive nuances articulated by the user community. The equation on text embedding $h_{p_i}^{txt}$ can be written as follows:
\begin{equation}
h_{p_i}^{txt}=BERT_{encoder}(X_{p_i}^{txt})    
\label{eq:BERT ENCODER}
\end{equation}

% \textbf{Text emebdding}

% We then want to give a unified emebdding of all users' review for each POI.  to learning the representation of the location from the category and short text review information from the user check-ins. The text information starts with lies in utilizing the order of check-in sequence from the users to the locations. We use this text information for the POI as the node features. 

\subsection{GOT Attention}
% \kaiqi{Connect to the previous sections. Is the GOT attention performed over GOT representation? How did you construct neighborhood graphs based on the contextual neighborhood? Or, perhaps we can just not mention graphs...otherwise, you need to explain the graph construction process..}

The GOT representation module produces unified representations, which the GOT attention mechanism integrates using a POI feature aggregator (Figure~\ref{fig:Architecture_MOEGOT}). Node features \( Q \), \( K \), and \( V \) are updated to \( q \), \( k \), and \( v \), enabling information aggregation from neighbors. The modified self-attention mechanism uses the dot-product of \( q \), \( k \), \( h^{occ} \), and \( h^{geo} \) to model pairwise relationships by incorporating spatial positions and co-occurrence. The final output \( z_i \) is obtained via element-wise product, concatenation, and linear transformation.

\subsubsection{Node feature aggregation}
% To fuse multi-context representations with multiple neighbors, we utilize GraphSage \cite{graphsage} as the fundamental aggregator. 
The geographical embeddings $h_{1,2}^{geo}$ and co-occurrence embeddings $h_{1,2}^{occ}$ are depicted by the edge from the first  $p_{1}$ to the second  $p_{2}$ within the GNNs, while the textual embeddings $h_{p_{1}^{txt}}$ and $h_{p_{2}^{txt}}$ serve as the node features for $p_{1}$ and $p_{2}$ respectively. For $p_{1}$, the aggregation of its node features involves summing geographical representations $h^{geo}$, co-occurrence representations $h^{occ}$, and the inherent textual representations $h^{txt}$ across all its neighbors. This aggregation is expressed as equation~\ref{eq:X_concat}:

\begin{equation}
\begin{aligned}
X_{p1}^{enh} = & Concat\left(\sum(h_{1,2}^{geo},h_{1,3}^{geo},\ldots,h_{1,n_{geo}}^{geo}), \right.\\
& \left. \sum(h_{1,2}^{occ},h_{1,3}^{occ},\ldots,h_{1,n_{occ}}^{occ}),h_{p1}^{txt} \right),
\label{eq:X_concat}
\end{aligned}
\end{equation}
where $X_{p1}^{enh}$ denotes the enhanced feature of $p_{1}$ after amalgamating textual, geographical, and co-occurrence information. $n_{geo}$ and $n_{occ}$ represent the number of geographical entities and co-occurrence entities in the aggregation, respectively, equivalent to the number of neighbors as predefined.

\subsubsection{Neighboring feature aggregation}

% Let $X^{enh}_{p}=[X^{enh}_{p_{1}:p_{n}}]$ represent the input embedding for all nodes following node feature aggregation. The process of aggregating neighboring features involves transforming the enhanced node embeddings $X_{p}^{enh}$ through a linear transformation to generate $Q$, $K$, and $V$ for the attention mechanism, which can be mathematically described as:
% \begin{equation}
% Q=X^{enh}_{p}W^{Q},
% \qquad
% K=X^{enh}_{p}W^{K},
% \qquad
% V=X^{enh}_{p}W^{V}
% \end{equation}
% where $W^{Q}$, $W^{K}$, and $W^{V}$ are parameter matrices for their respective transformations.

% In order to capture the local graph structure surrounding each node and learn the significance of neighboring nodes through edge-specific selections of $q$, $k$, $v$ based on edge indices, we transform $Q$, $K$, and $V$ into $q$, $k$, and $v$ as follows:
% \begin{equation}
%     q = Q[\text{edge}[0, :]], 
%     \qquad
%     k = K[\text{edge}[1, :]], 
%     \qquad 
%     v = V[\text{edge}[1, :]]
% \end{equation}
 
% This transition to edge-level queries, keys, and values enables the model to dynamically prioritize information flow between connected nodes. By leveraging both node features and edge attributes, the model effectively aggregates information from neighbors, capturing local structural and relational properties in a unified manner.

Let \( X^{enh}_{p} = [X^{enh}_{p_{1}:p_{n}}] \) denote the input embeddings for all nodes after feature aggregation. Neighboring feature aggregation involves transforming \( X^{enh}_{p} \) into query, key, and value matrices (\( Q, K, V \)) and further indexing them based on edge connectivity to produce edge-specific representations (\( q, k, v \)):
\begin{equation}
\begin{aligned}
Q &= X^{enh}_{p}W^{Q}, \quad K = X^{enh}_{p}W^{K}, \quad V = X^{enh}_{p}W^{V}, \\
q &= Q[\text{edge}[0, :]], \quad k = K[\text{edge}[1, :]], \quad v = V[\text{edge}[1, :]],
\end{aligned}
\end{equation}

where \( W^{Q} \), \( W^{K} \), and \( W^{V} \) are learnable parameter matrices.
The edge-level transformation enables the model to prioritize information flow between connected nodes, effectively integrating node features and edge attributes to capture local graph structures.

\subsubsection{Modified attention mechanism}
We propose a relative position encoder to simultaneously include both geographical and co-occurrence embeddings. It is designed to compute the output $z_{i}$ for the $i$-th POI by integrating geographical and co-occurrence contexts into the attention scores, as expressed by Equation \ref{eq:Mod_attention}:

\begin{equation}
\begin{aligned}
\boldsymbol{z}_{i} &= \sum_{j=1}^{n} \alpha_{i j} \boldsymbol{v}_{j} \in \mathbb{R}^{d_{k}}, \quad \alpha_{i j} = \frac{\exp \left(e_{i j}\right)}{\sum_{k=1}^{n} \exp \left(e_{i k}\right)}, \\
e_{ij} &= \frac{\operatorname{sum}\left(\boldsymbol{q}_{i} \odot \boldsymbol{k}_{j} \odot \boldsymbol{h}^{geo}_{i j} \odot \boldsymbol{h}^{occ}_{i j}\right)}{\sqrt{d_{k}}},
\end{aligned} 
\label{eq:Mod_attention}
\end{equation}
where $\alpha_{i j}$ represents the normalized weight using a softmax function, and $e_{i j}$ denotes the attention score from node $j$ to $i$. 

Subsequently, we concatenate all outputs from different heads through a projection operation, which consolidates and refines the information captured by each head into a cohesive representation, as demonstrated in the equation below:

\begin{equation}
    Z_i = Concat(\boldsymbol{z}^{(1)}_i,\boldsymbol{z}^{(2)}_2,\ldots,\boldsymbol{z}_{i}^{(h)})W^O,
\end{equation}
where $W^O$ is the parameter matrix for linear transformation and $\boldsymbol{z}_{i}^{(h)}$ is the output of the $h$-th head.

Furthermore, to enhance the model's generalization capabilities and mitigate overfitting, we apply normalization and dropout techniques to this projected representation to generate the final representation  $z_{i}^{final}$.

\begin{equation}
    z_{i}^{final} = Droupout(LayerNorm(Z_i))
\end{equation}

\subsection{Adaptive Representation
Aggregator}

%The final POI representation under each subgraph is denoted as $z_{i,o}^{final}$, where $o \in [1:4]$ denotes the subgraph set.

Following the aggregation of neighborhoods using the GOT attention mechanism, the adaptive representation aggregator is applied to dynamically select the expert associated with each contextual graph motivated by MoE~\cite{moe}.
Unlike existing pre-trained model that produces a fixed representation for each POI~\cite{SpaBERT},  the adaptive representation learning technique can dynamically adjust the representation to different tasks without retraining.
The subgraph representations can be further minimized by Jensen-Shannon (JS) divergence on subgraphs and original graphs. The structure can be delineated by:

\begin{equation}
h_{i}^{\prime} = \sigma\left(\sum_{o=1}^{4} G\left(X_{p_{i}}^{enh}\right)_{o} \cdot z_{i,o}^{final}\right),
\end{equation}

\begin{equation}
Q(z_{i,o}^{final}) = W_{g} \cdot z_{i,o}^{final} + \epsilon \cdot \text{Softplus}(z_{i,o}^{final} \cdot W_{n}),
\end{equation}
\begin{equation}
G(h_{i}) = \text{softmax}(\text{Topk}(Q(z_{i,o}^{final}))),
\end{equation}
where $h_{i}^{\prime}$ represents the latent representation of MoE, $o$ denotes the $o$-th expert, $G\left(X_{p_{i}}^{enh}\right)_{o}$ stands for the gating weights for expert $o$, and $\epsilon \in \mathcal{N}(0,1)$ is the standard Gaussian noise. $W_{g} \in \mathbb{R}^{s \times n}$ and $W_{n} \in \mathbb{R}^{s \times n}$ are learnable weights that determine the clean and noisy scores, respectively. Finally, $G(h_{i})$ produces a sparse probability distribution over experts, decides how strongly each expert's output contributes to the framework.

In addition to dynamically assigning experts to contextual subgraphs, the adaptive representation aggregator refines subgraph topology by minimising the JS divergence between subgraph and original graph feature distributions, thereby preserving global structural fidelity. The JS divergence is defined as:

\begin{equation}
D_\text{JS}(P(G_s) \| P(G)) = \frac{1}{2} D_\text{KL}(P(G_s) \| M) + \frac{1}{2} D_\text{KL}(P(G) \| M),  
\label{eq:JSdiv}
\end{equation}
where \( M = \frac{1}{2}(P(G_g) + P(G)) \) is the midpoint distribution in equation~\ref{eq:JSdiv}, and \( D_\text{KL} \) is the Kullback-Leibler divergence. 

The context-specific subgraph $G_s$ is not statically defined but is dynamically constructed through a MoE mechanism. Rather than selecting a single predefined topology from the candidate set $G_g$ as Figure~\ref{fig:Architecture_MOEGOT} shows, which comprises subgraphs initialised using four sampling strategies, $G_s$ is instead formed as a weighted combination of these alternatives.
 The MoE gating function $G(h_i)$ adaptively assigns importance to each subgraph, enabling task-aware, localised modelling while preserving structural diversity and alignment with global graph semantics.

The MoE adaptively selects subgraph topologies by learning an optimal weighting scheme \( G(h_i) \) to minimize JS divergence. This ensures subgraph representations align with the global graph's structure and attributes, effectively retaining local and global information to enhance POI embedding robustness and expressiveness.

\subsection{Encoder-Decoder Structure}
Ultimately, within the Encoder-Decoder architecture, an MLP-based decoder is utilized to interpret and extract latent representation provided by the adaptive representation learning module.

% \kaiqi{The following paragraph is not very useful. You should have mentioned these before. This Section 3.6 can be about training. You can start with a brief introduction to the decoder and say we use MLM approach to train the embeddings (merge 3.6.1 and 3.6.2). Then, will talk about the loss functions.}

% To facilitate the integration of inter-relational information among POIs, we designed a directed graph. This graph incorporates node features derived from text features extracted by our BERT encoder, denoted as $h^{txt}$, and edge features comprising both Co-occurrence ($h^{occ}$) and Geographical ($h^{geo}$) representations with directional attributes as detailed in the preceding section. Acknowledging the impact of location, we proceed to generate an informative embedding for each location by aggregating edge features across all neighboring nodes. This aggregation process leads to each node accumulating the sum of edge features from its neighboring nodes. Subsequently, these aggregated edge features are concatenated with the original node feature $X_{pi}^{txt}$ to yield an enhanced node feature representation ($X_{pi}^{enh}$). Utilizing a modified attention mechanism employing multi-head attention, we aggregate information from neighboring nodes to derive the final representation ($z_{i}^{final}$) for POIs.

\subsubsection{Decoder}

% We adopt the simple Multi-Layer Perceptron (MLP) as the decoder to  compell the encoder to capture the most relevant features in a more condensed form \cite{ma2022pre}. This is particularly crucial in models where the quality of learned embeddings directly influences performance on downstream tasks. 

We utilize three distinct MLPs as $FFN_{txt}$, $FFN_{geo}$ and $FFN_{occ}$ for decoding the Co-occurrence representation, Geographical representation, and Mixed user text representation\cite{CTLE}.

Inspired by BERT in the NLP models, we implement the Masked Language Model (MLM) to construct the self-supervised learning fashion in the training phase. As figure~\ref{fig:Architecture_MOEGOT} shows, we randomly mask 20\% records of $X_{p_{i}}^{txt}$, $S_{p_{i}}^{txt}$ and replace the corresponding text with special token 
$[X_{p_{i}}^{txt}]_{m}$, $[S_{p_{i}}^{txt}]_{m}$ respectively. Therefore, the $[e^{occ}_{i,n}]_{m}$, $[e^{geo}_{i,n}]_{m}$ will be represented accordingly by GOT representation following ~\ref{sec:GOT representation}. Let $f$ be the mapping function in the encoder, and we aim to predict the masked token formulated by:

\begin{equation}
\begin{aligned}
\hat{X}_{p_{i}}^{txt} &= FFN_{txt}(f([X_{p_{i}}^{txt}]_{m})) \quad &\hat{e}_{i,n}^{occ} &= FFN_{occ}(f([e_{i,n}^{occ}]_{m})) \\
\hat{e}_{i,n}^{geo} &= FFN_{geo}(f([e_{i,n}^{geo}]_{m}))
\end{aligned}
\label{eq:training_process}
\end{equation}

\subsubsection{Loss Function}

As our methodology is designed to be task-agnostic, emphasizing the acquisition of general representations applicable across diverse applications, the loss function is tailored not for task-specific objectives but rather for reconstructing or closely approximating the original input information, spanning text, spatial, and co-occurrence data. Therefore, we utilize a weighted loss strategy:

% During the training phase, the model and its decoders are optimized to minimize the discrepancy between predicted and true features. Individual losses are computed for each feature type, and a weighted loss strategy is employed to balance their contributions. This approach enables the model to learn a comprehensive embedding that encapsulates all aspects of POI information.
% This ensures that our model effectively focuses on encoding and decoding the fundamental characteristics of POIs and their relationships.

%The decoder for the text is simply to generate the text information from the $z_{i}^{final}$. The decoder for the Geographical information is aiming to decode the spatial relationship between POI pairs. Therefore we take two POI pairs $p_i$ and $p_j$, and concatenate their embeddings ($z_{i}^{final}$, $z_{j}^{final}$) to decode spatial features distance and azimuth back. Similarly, for co-occurrence representation, it decodes the frequency of occurrence between POI pairs from their concatenated final embeddings.

% During the training phase, the model and its decoders are optimized to minimize the discrepancy between predicted and true features. Individual losses are computed for each feature type, and a weighted loss strategy is employed to balance their contributions. This approach enables the model to learn a comprehensive embedding that encapsulates all aspects of POI information.

\noindent\textbf{Textual Loss} ($L_{txt}$) measures the discrepancy between predicted and true text features, using binary cross-entropy for one-hot encoded text data:
\begin{equation}
L_{i}^{txt}=CrossEntropy(\hat{X}_{p_{i}}^{txt},X^{txt}_i)
\end{equation}

\noindent\textbf{Geographical Loss} ($L_{geo}$) gauges the discrepancy in predicting spatial features (distance and azimuth) between pairs of Points of Interest (POIs), utilizing mean squared error (MSE):

\begin{equation}
    L_{ij}^{geo} = MSE (\hat{e}_{i,n}^{geo},e^{geo}_{ij})
\end{equation}

\noindent\textbf{Co-occurrence Loss} ($L_{occ}$) assesses the error, i.e., MSE, in predicting the frequency of co-occurrence between POI pairs:
\begin{equation}
    L_{ij}^{occ} = MSE (\hat{e}_{i,n}^{occ},e^{occ}_{ij})
\end{equation}

% \noindent\textbf{MOE Loss}
% Considering scenarios where a single expert consistently gains favor, a phenomenon stemming from the self-reinforcing nature of imbalance where certain experts proliferate more rapidly than others, thus increasing their frequency of selection, we introduce an importance loss mechanism to offset the impact of multiple contexts \cite{GMOE,SparseMOE}:

% \begin{equation}
% \text { Imp}(H)=\sum_{h_{i} \in H, g \in G\left(h_{i}\right)} g, L_{\text {imp }}(H)=C V(\text {Imp}(H))^{2},    
% \end{equation}
% where the importance score $\text{Imp}(H)$ represents the sum of each node's gate value $g$ across the entire batch. CV denotes the coefficient of variation. The importance loss $L_{\text {imp }}(H)$ ensures that important experts are selected. Except for the importance score, JS divergence provides another loss function in the training.

\noindent\textbf{MOE Loss}  
To address the imbalance caused by certain experts being over-selected due to self-reinforcement, we introduce an importance loss mechanism to balance multiple contexts \cite{GMOE,SparseMOE}:
\begin{equation}
\text{Imp}(H)=\sum_{h_{i} \in H, g \in G(h_{i})} g, \quad L_{\text{imp}}(H)=C V(\text{Imp}(H))^{2},
\end{equation}
where \(\text{Imp}(H)\) is the sum of each node's gate value \(g\) across the batch, and \(CV\) is the coefficient of variation. This importance loss \(L_{\text{imp}}(H)\) ensures balanced expert selection. Additionally, JS divergence serves as an auxiliary loss during training.

\noindent\textbf{Overall Loss}
To allow the model to automatically balance multiple objectives, we define the overall loss ($L_{\text{total}}$) as a softmax-normalised weighted sum of individual loss components:

\begin{equation}
    L_{\text{total}} = \sum_{k} \alpha_k L_k, \quad \text{where} \quad \alpha_k = \frac{\exp(w_k)}{\sum_{j} \exp(w_j)}
\end{equation}

Here, $\{w_k\}$ denotes a set of learnable scalar parameters associated with each loss component $L_k \in \{L_{\text{txt}}, L_{\text{geo}}, L_{\text{occ}}, L_{\text{imp}}, L_{\text{JS}}\}$.
% The softmax function ensures that the weights $\alpha_k$ form a valid probability distribution (i.e., $\alpha_k \in [0,1]$ and $\sum_k \alpha_k = 1$), thereby avoiding degenerate solutions and enabling stable training. 
This learnable weighting scheme allows the model to dynamically adapt the contribution of each context during training process.

\noindent\textbf{Time complexity}
 The time complexity can be represented as: $O(k \sum_{h_i \in H} \sum_{j \in N_i}(|V|FF^{\prime}+|E|F^{\prime}))$, 
% \begin{equation}
% \begin{aligned}
% &O((|V|FF^{\prime}+|E|F^{\prime})(\sum_{i}G(h_{i})_{o}))\\ 
% &G\left(h_{i}\right)_{o}=\left\{\begin{array}{ll}
% 0 & \text { if } G\left(h_{i}\right)_{o}=0 \\
% 1 & \text { otherwise }
% \end{array}
% \end{aligned}
% \end{equation}
where the right part represents the time complexity from attention\cite{gat}, where $F, F^{\prime}$ is the number of input features and output features, $|V|$ and $|E|$ represent the number of nodes and edges respectively. The left part denotes the time complexity of fusing $o$ experts, similar as~\cite{GMOE}, the $h_{i}$ denotes the input feature of $i$-th node, $j$ denotes the $j$-th node; $H$ and $N_{i}$ denote the number of input feature dimension and neighborhood of $i$-th node respectively.
\vspace{8pt} % Slightly reduces the space, adjust as needed
\section{Experiments}

\subsection{EXPERIMENTAL SETUP}

% \subsubsection{Research questions}
% We conducted experiments to assess the performance of \textsf{MOEGOT} compared to the state-of-the-art POI representation learning method SpaBERT \cite{SpaBERT} across various POI downstream tasks. Our investigation aimed to address four key questions:
% \begin{itemize}[leftmargin=*]
% \item 
% \textbf{RQ1:} Does \textsf{MOEGOT} outperform the baseline pre-trained models across different downstream tasks? 
%   \item 
% \textbf{RQ2:} What is the individual effectiveness of each component within \textsf{MOEGOT}?
% \item 
% \textbf{RQ3:} What is the effectiveness of MOEGOT in multiple POI contexts?
% \item
% \textbf{RQ4:} How sensitive is MOEGOT under different hyperparameter settings?
% \item 
% \textbf{RQ5:} Can MOEGOT perform well on other task?
  % \item 
% \textbf{RQ5:} Is TransGOT efficient under different POI tasks? How well does TransGOT perform w.r.t the data size and the number of neighbours in the network?
% \end{itemize}

\subsubsection{Datasets}

\begin{table}[ht]
\vspace{-1em}
\centering
\caption{Statistics of Datasets}
\label{tab:dataset}
\renewcommand{\arraystretch}{0.8} % Reduce row spacing
\setlength{\tabcolsep}{0.8mm} % Reduce column spacing
%\scriptsize % Use smaller font size
\begin{tabular}{cclccc}
\toprule
Dataset & \#Category & City/State & \#User & \#POI & \#Check-in \\ \midrule
\multirow{2}{*}{Foursquare} & \multirow{2}{*}{9} & New York & 1,083 & 9,989 & 179,468 \\
& & Tokyo & 2,293 & 15,177 & 494,807 \\ \hdashline
\multirow{2}{*}{Yelp} & \multirow{2}{*}{21} & Louisiana & 40,183 & 7,811 & 106,469 \\
& & Nevada & 18,573 & 5,317 & 53,053 \\ 
\bottomrule
\end{tabular}
\vspace{-1em}
\end{table}

\textbf{Foursquare}\footnote{https://sites.google.com/site/yangdingqi/home/foursquare-dataset} comprises check-in data from two major cities, New York (NY) and Tokyo City, spanning from April 2012 to February 2013. Each check-in entry includes geographical coordinates, timestamps, and the category of the visited POI. \textbf{Yelp}\footnote{https://www.kaggle.com/datasets/yelp-dataset/yelp-dataset} is a subset of Yelp's businesses, reviews, and users' check-in data. We narrowed down the dataset to focus on information from the states of Nevada (NV) and Louisiana (LA).

% In both datasets, geolocation and category information are included. Foursquare contains user check-ins, while Yelp uses tips as check-ins, which also include short text messages that users leave for each location. During the pre-processing stages, we excluded users with fewer than five check-ins and POI locations that were checked-in less than five times. We considered the temporal dimension of check-ins when generating user trajectories. Using this approach, we segmented the check-in sequence of a user if there was a time gap between two consecutive POIs in their check-in history greater than a fixed amount of time as discussed in Section \ref{sec:init}. We set this fixed amount of time to 48 hours. It is likely that a user activity phase characterized by a gap of more than 48 hours between check-ins indicates a distinct activity phase, and treating such check-ins as part of the same trajectory could distort the true movement patterns and preferences of that user.

\subsubsection{Evaluation Metrics} We calculate two widely used measure in recommender systems: Recall@K metric and NDCG@K~\cite{NAVIP}, we only show Recall@K by simplicity on this paper\footnote{Extensive experiments on NDCG@k: https://anonymous.4open.science/r/POIemb-
F161/README.}. They signify the proportion of relevant items within the top K items compared to all relevant items:

% \begin{equation}
%     Recall@K = \frac{\sum_{i=1}^{N} | \text{Hit}(i) |}{N}
% \end{equation}
% Where $i$ is the $i$-th test data point and $N$ denotes the total number of test data points.

% ,$Rank(i)$ is the ranking of the real next hop position in the $i$-th test data in the set of the top K locations in the prediction result of the $i$-th test data

\subsubsection{Baseline}
 The baselines of pre-trained embedding models:  \textbf{FC}: fully connected model. \textbf{Rand}: Random initialization. \textbf{CTLE}~\cite{CTLE}: A pre-trained
method considering contextual neighbors within trajectories.
The baselines of next POI and next category recommendation: \textbf{GETNEXT}~\cite{Getnext},  \textbf{FPMC}~\cite{FPMC}, \textbf{SERM}~\cite{serm}, \textbf{HSTLSTM}~\cite{hstlstm}, \textbf{LSTPM}~\cite{LSTPM}. The baselines for POI Recommendation: \textbf{EEDN}~\cite{EEDN}
\textbf{VanillaMF}~\cite{vanillaMF}
\textbf{RankNet}~\cite{rankNet}
\textbf{NGCF}~\cite{ngcf}
\textbf{NeuMF}~\cite{neuMF}.

\subsubsection{Implementation Details}
The number of heads $A$ is set to 2 in \textsf{AdaptGOT}, neighbors $k$ is set to 5, layer normalization epsilon ($eps$) is set to $1\mathrm{e}{-12}$, the encoder dimensions $d_{k}$ is 16 and a dropout probability ($p$) is set of 0.1. We use Adam optimizer\cite{Adam} with a learning rate ($l_{r}$) of 0.001 and epoch of 100. For SpaBERT \cite{SpaBERT}, the distance factor $Z$ is 0.0001, and the distance separator $D_{SEP}$ is set to 20. The embedding size $d_{model}$ is set to 768. 

The hyper-parameter is tuned following the grid search,
For \textsf{AdaptGOT}: $A \in [1,2,4]$, $k \in [1:5]$, $d_{k} \in [4,8,16,32]$; For others: $Z \in[0.001, 0.0001, 0.00001]$, $d_{model} \in[256,512,768]$. Our experiments were conducted on an AMD Ryzen 7 7700X CPU, 64GB RAM and an NVIDIA RTX 4090 GPU.

% For the downstream tasks stage, the baseline models are utilised by the source code provided by the authors. We replace the baseline models' POI embeddings with our pre-trained embeddings for all downstream tasks, comparing them against the original randomly generated embeddings. All the rest just follow the original paper approaches. 

% For the next POI category prediction, we trained the training set with the category information during the pre-training and use the embedding to predict the category of the next POI. We map all the lower categories to the top-level category. 

% For POI recommendations, we pre-processed the dataset into a set of POIs within a grid, dividing coordinates by a region rate of 0.008 and using the sum of pre-trained POI embeddings as the region embeddings for comparison. 

\subsection{Overall Comparison}
We report the results on three downstream tasks in Table~\ref{tab:Threetasks}.

\begin{table*}[htbp]
\vspace{-1em}
\centering
\Large % Reduce font size
\setlength{\tabcolsep}{2pt} % Adjust column spacing
\renewcommand{\arraystretch}{1.2} % Adjust row spacing
\caption{Overall comparison between AdaptGOT (GOT) and other pre-training model (PM) on three downstream models (DM).}
\label{table:poi_recommendation}
\resizebox{\textwidth}{!}{ % Adjust table width to fit
\renewcommand{\arraystretch}{0.9} 
\begin{tabular}{
c|c|ccc|ccc|ccc|ccc|c|ccc|ccc % Removed first and last vertical lines
} 
\hline
\multicolumn{2}{c|}{Dataset and tasks}         & \multicolumn{3}{c!{\vrule width 0.3pt}}{NY-Next POI rec}                                                                            & \multicolumn{3}{c|}{LA-Next POI rec}                                                                            & \multicolumn{3}{c!{\vrule width 0.3pt}}{NY-Next cate rec}                                                                           & \multicolumn{3}{c|}{LA-Next cate rec}                                                                           & Baseline                     & \multicolumn{3}{c!{\vrule width 0.3pt}}{NY-POI rec}                                                                                 & \multicolumn{3}{c}{LA-POI rec}                                                                                 \\ \hline
DM                        & PM   & Rec@5 & Rec@10 & Rec@15 & Rec@5 & Rec@10 & Rec@15 & Rec@5 & Rec@10 & Rec@15 & Rec@5 & Rec@10 & Rec@15 & DM                      & Rec@5 & Rec@10 & Rec@15 & Rec@5 & Rec@10 & Rec@15 \\ \hline
 & Rand &  0.283 & 0.394  & 0.461  & 0.072 & 0.087  & 0.101  & 0.669 & 0.820  & 0.890  & 0.536 & 0.739  & 0.831  &                         & 0.104 & 0.175  & 0.257  & 0.143 & 0.196  & 0.234  \\
& FC   & 0.277 & 0.382  & 0.446  & 0.067 & 0.082  & 0.089  & 0.678 & 0.857  & 0.911  & 0.548 & 0.755  & 0.842  &                         & 0.107 & 0.179  & 0.262  & 0.151 & 0.207  & 0.238  \\
& CTLE & 0.252 & 0.364  & 0.431  & 0.058 & 0.076  & 0.078  & 0.668 & 0.826  & 0.855  & 0.541 & 0.722  & 0.837  &                         & 0.112 & 0.173  & 0.257  & 0.146 & 0.198  & 0.232  \\
 & Spa  & 0.245 & 0.347  & 0.405  & 0.043 & 0.063  & 0.063  & 0.613 & 0.771  & 0.857  & 0.531 & 0.763  & 0.845  &                         & 0.096 & 0.169  & 0.252  & 0.108 & 0.130  & 0.141  \\
\multirow{-5}{*}{FPMC}    & GOT  & \textbf{0.312} & \textbf{0.423} & \textbf{0.486} & \textbf{0.086} & \textbf{0.125} & \textbf{0.156} & \textbf{0.711} & \textbf{0.862} & \textbf{0.927} & \textbf{0.605} & \textbf{0.794} & \textbf{0.885} & \multirow{-5}{*}{VaMF} & \textbf{0.121} & \textbf{0.186} & \textbf{0.273} & \textbf{0.163} & \textbf{0.224} & \textbf{0.269} \\ \hline & Rand & 0.399 & 0.485  & 0.529  & 0.064 & 0.079  & 0.089  & 0.759 & 0.874  & 0.928  & 0.639 & 0.782  & 0.866  &                         & 0.089 & 0.143  & 0.201  & 0.175 & 0.225  & 0.260  \\ & FC   & 0.386 & 0.473  & 0.521  & 0.058 & 0.071  & 0.075  & 0.747 & 0.853  & 0.915  & 0.618 & 0.753  & 0.847  &                         & 0.092 & 0.147  & 0.208  & \textbf{0.182} & 0.231  & 0.268  \\& CTLE & 0.388 & 0.476  & 0.525  & 0.063 & 0.075  & 0.079  & 0.756 & 0.874  & 0.914  & 0.559 & 0.736  & 0.789  &                         & 0.097 & 0.157  & 0.219  & 0.178 & \textbf{0.237} & \textbf{0.269} \\& Spa  & 0.395 & 0.483  & 0.528  & \textbf{0.078} & 0.084  & 0.089  & 0.759 & 0.883  & 0.929  & 0.564 & 0.743  & 0.802  &                         & 0.092 & 0.149  & 0.212  & 0.173 & 0.229  & 0.260  \\

\multirow{-5}{*}{LSTPM}   & GOT  & \textbf{0.426} & \textbf{0.513} & \textbf{0.557} & \textbf{0.078} & \textbf{0.112} & \textbf{0.121} & \textbf{0.789} & \textbf{0.913} & \textbf{0.957} & \textbf{0.662} & \textbf{0.824} & \textbf{0.884} & \multirow{-5}{*}{RkNet} & \textbf{0.109} & \textbf{0.187} & \textbf{0.249} & 0.175 & 0.229  & 0.263  \\ \hline
& Rand & 0.327 & 0.392  & 0.428  & 0.033 & 0.059  & 0.069  & 0.705 & 0.840  & 0.901  & 0.580 & 0.715  & 0.784  &                         & 0.095 & 0.175  & 0.239  & 0.128 & 0.205  & 0.248  \\
& FC   & 0.339 & 0.407  & 0.415  & 0.038 & 0.044  & 0.047  & 0.701 & 0.827  & 0.896  & 0.568 & 0.707  & 0.776  &                         & 0.097 & 0.173  & 0.237  & 0.125 & 0.201  & 0.243  \\
 & CTLE & 0.315 & 0.384  & 0.418  & 0.057 & 0.079  & 0.083  & 0.708 & 0.836  & 0.905  & 0.577 & 0.713  & 0.784  &                         & 0.094 & 0.178  & 0.231  & 0.117 & 0.198  & 0.245  \\
& Spa  & 0.309 & 0.376  & 0.411  & 0.051 & \textbf{0.071} & 0.076  & 0.699 & 0.838  & 0.899  & 0.557 & 0.814  & 0.880  &                         & \textbf{0.111} & 0.170  & 0.242  & \textbf{0.153} & \textbf{0.213} & \textbf{0.262} \\
\multirow{-5}{*}{HST}     & GOT  & \textbf{0.358} & \textbf{0.431} & \textbf{0.470} & \textbf{0.048} & \textbf{0.071} & \textbf{0.087} & \textbf{0.724} & \textbf{0.849} & \textbf{0.904} & \textbf{0.606} & \textbf{0.748} & \textbf{0.791} & \multirow{-5}{*}{NGCF} & 0.093 & \textbf{0.187} & \textbf{0.254} & 0.147 & 0.201  & 0.259  \\ \hline
& Rand & 0.397 & 0.495  & 0.544  & 0.062 & 0.078  & 0.102  & 0.727 & 0.852  & 0.907  & 0.581 & 0.776  & 0.882  &                         & 0.101 & 0.166  & 0.254  & 0.142 & 0.208  & 0.254  \\
 & FC   & 0.394 & 0.489  & 0.538  & 0.057 & 0.072  & 0.094  & 0.732 & 0.859  & 0.912  & 0.592 & 0.792  & 0.831  &                         & 0.093 & 0.153  & 0.238  & 0.128 & 0.185  & 0.241  \\
& CTLE & 0.394 & 0.492  & 0.544  & 0.061 & 0.092  & 0.112  & 0.738 & 0.862  & 0.917  & 0.599 & 0.788  & 0.844  &                         & 0.091 & 0.148  & 0.234  & 0.122 & 0.178  & 0.233  \\
& Spa  & 0.399 & 0.495  & 0.551  & 0.065 & 0.096  & 0.121  & 0.754 & 0.877  & 0.926  & \textbf{0.624} & 0.811  & 0.857  &                         & \textbf{0.095} & 0.179  & 0.233  & \textbf{0.153} & 0.203  & 0.250  \\
\multirow{-5}{*}{SERM}    & GOT  & \textbf{0.427} & \textbf{0.523} & \textbf{0.561} & \textbf{0.084} & \textbf{0.112} & \textbf{0.140} & \textbf{0.758} & \textbf{0.880} & \textbf{0.928} & 0.603 & \textbf{0.817} & \textbf{0.873} & \multirow{-5}{*}{NeuMF} & \textbf{0.095} & \textbf{0.180} & \textbf{0.255} & \textbf{0.153} & \textbf{0.209} & \textbf{0.256} \\ \hline
 & Rand & 0.416 & 0.508  & 0.554  & 0.079 & 0.104  & 0.128  & 0.747 & 0.865  & 0.903  & 0.592 & 0.794  & 0.850  &                         & 0.117 & 0.179  & 0.268  & 0.158 & 0.217  & 0.268  \\
 & FC   & 0.423 & 0.513  & 0.561  & 0.082 & 0.108  & 0.135  & 0.752 & 0.871  & 0.912  & 0.607 & 0.788  & 0.839  &                         & 0.119 & 0.183  & 0.271  & 0.166 & 0.228  & 0.271  \\
 & CTLE & 0.432 & 0.525  & 0.564  & 0.088 & 0.115  & 0.139  & 0.763 & 0.878  & 0.916  & 0.623 & 0.811  & 0.845  &                         & 0.126 & 0.190  & 0.277  & 0.165 & 0.231  & 0.277  \\
& Spa  & 0.441 & 0.527  & 0.589  & 0.096 & 0.119  & 0.143  & 0.771 & 0.886  & 0.932  & 0.631 & 0.827  & 0.864  &                         & 0.135 & 0.195  & 0.293  & 0.171 & 0.235  & 0.286  \\
\multirow{-5}{*}{GETNEXT} & GOT  & \textbf{0.447} & \textbf{0.536} & \textbf{0.593} & \textbf{0.102} & \textbf{0.126} & \textbf{0.152} & \textbf{0.775} & \textbf{0.894} & \textbf{0.946} & \textbf{0.648} & \textbf{0.835} & \textbf{0.892} & \multirow{-5}{*}{EEDN} & \textbf{0.143} & \textbf{0.196} & \textbf{0.297} & \textbf{0.177} & \textbf{0.246} & \textbf{0.293} \\ \hline
\end{tabular}
}
\label{tab:Threetasks}
\end{table*}

\subsubsection{Next POI Recommendation}
\label{sec:NEXTPOIREC}

The table demonstrates GOT's consistent superiority over baseline models, excelling in synthesizing contextual and sequential information for next POI recommendations. AdaptGOT achieves the highest recall scores, particularly in dense, multi-context environments like New York, where it improves Rec@5 by 23.8\% over SpaBERT under FPMC and 15.5\% under GETNEXT. In Los Angeles, GOT shows a 48.3\% Rec@5 improvement over CTLE under FPMC and 36.5\% over SpaBERT. AdaptGOT also leads in tasks like LSTPM and HST, achieving substantial gains such as 33.3\% in Rec@10 over CTLE under LSTPM. In contrast, models like SpaBERT and CTLE fall short in capturing fine-grained contextual information. 

\subsubsection{Next Category Prediction}

% GOT demonstrates consistent and significant improvements over SpaBERT and CTLE across all models and metrics, particularly in the NY-Next dataset. For instance, in FPMC, GOT improves recall@5 by 15.9\% compared to SpaBERT and by 9.8\% compared to CTLE, showcasing its superior ability to synthesise context and check-in sequences. Similar trends are observed in GETNEXT, where GOT achieves up to 4.0\% and 2.7\% higher recall@5 over CTLE and SpaBERT, respectively, in the LA-Next dataset. Overall, the results underline \textsf{MOEGOT}'s strong contextual integration capabilities, which prove especially advantageous in NY and LA datasets.

GOT shows significant improvements over SpaBERT and CTLE across all models and metrics, particularly in the NY-Next dataset. In FPMC, it improves Rec@5 by 15.9\% over SpaBERT and 9.8\% over CTLE, highlighting its ability to synthesise context and check-in sequences. In GETNEXT on the LA dataset, GOT achieves 4.0\% and 2.7\% higher Rec@5 over CTLE and SpaBERT, respectively. These results underscore \textsf{AdaptGOT}'s strong contextual integration, especially in NY and LA.

\subsubsection{POI Recommendation}

% MOEGOT consistently outperforms baseline models across POI recommendation tasks. In New York, MOEGOT achieves significant improvements, including a 16.7\% increase in Rec@5 and 20.8\% in Rec@15 under FPMC~\cite{FPMC} compared to CTLE~\cite{CTLE}, and a 13.3\% higher Rec@5 in GETNEXT over SpaBERT. Similarly, MOEGOT demonstrates a 44.5\% increase in Rec@5 over SpaBERT and a 14.3\% improvement in Rec@15 compared to CTLE under FPMC on LA. Across LSTPM and HST, GOT consistently leads with notable gains, such as a 14.6\% improvement in Rec@5 over SpaBERT in LA under LSTPM and a 15.4\% increase in Rec@5 over CTLE under HST. These results highlight MOEGOT's robust ability to integrate multiple contexts, achieving superior POI prediction accuracy across diverse baselines and datasets.

AdaptGOT outperforms baselines in POI recommendation. In New York, it improves Rec@5 by 16.7\% and Rec@15 by 20.8\% under FPMC compared to CTLE and achieves a 13.3\% higher Rec@5 in GETNEXT over SpaBERT. In LA, it shows a 44.5\% gain in Rec@5 over SpaBERT and a 14.3\% improvement in Rec@15 compared to CTLE under FPMC. AdaptGOT also achieves notable gains in Rec@5, including 14.6\% over SpaBERT under LSTPM and 15.4\% over CTLE under HST, highlighting its superior ability to integrate multiple contexts for accurate POI predictions.
\subsection{Ablation Studies}

\subsubsection{Sampling and GOT representation}
\label{sec:mixsamp}

To evaluate the influence of each component within our representation on model performance, we conducted experiments on the next POI recommendation task using SERM \cite{serm} in the Foursquare (NY) dataset. Here
are the experimental settings:
\begin{itemize}[leftmargin=*]
    \item w/o Cate, Imp, KNN and Density: it removes the category-aware sampling, importance sampling, KNN sampling, and density-based sampling respectively;
    \item w/o Occ, Txt, and Learn: it removes the co-occurrence representation and the text representation and freezes the POI embedding learning in the downstream training phase.
\end{itemize}

\begin{figure}[h]
\vspace{-1em}
\centering
\subfigure[Mixed sampling by foursquare]
{
\begin{minipage}[t]{0.48\linewidth}
\centering
\includegraphics[scale=0.26]{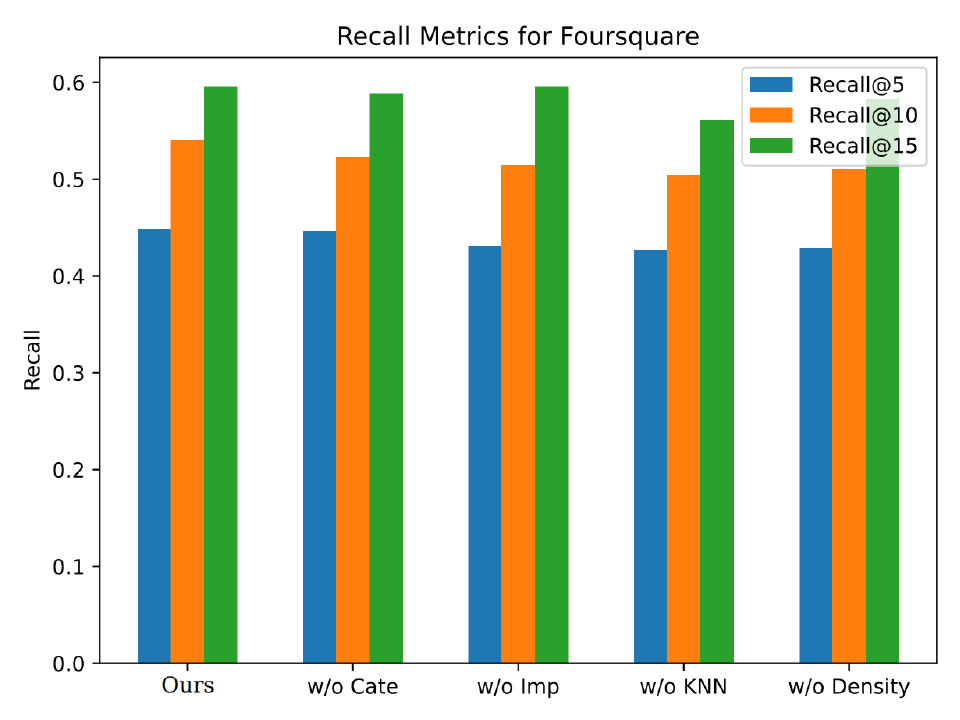}
%\caption{fig2}
\end{minipage}}
\subfigure[Mixed sampling by Yelp]
{\begin{minipage}[t]{0.48\linewidth}
\centering
\includegraphics[scale=0.26]{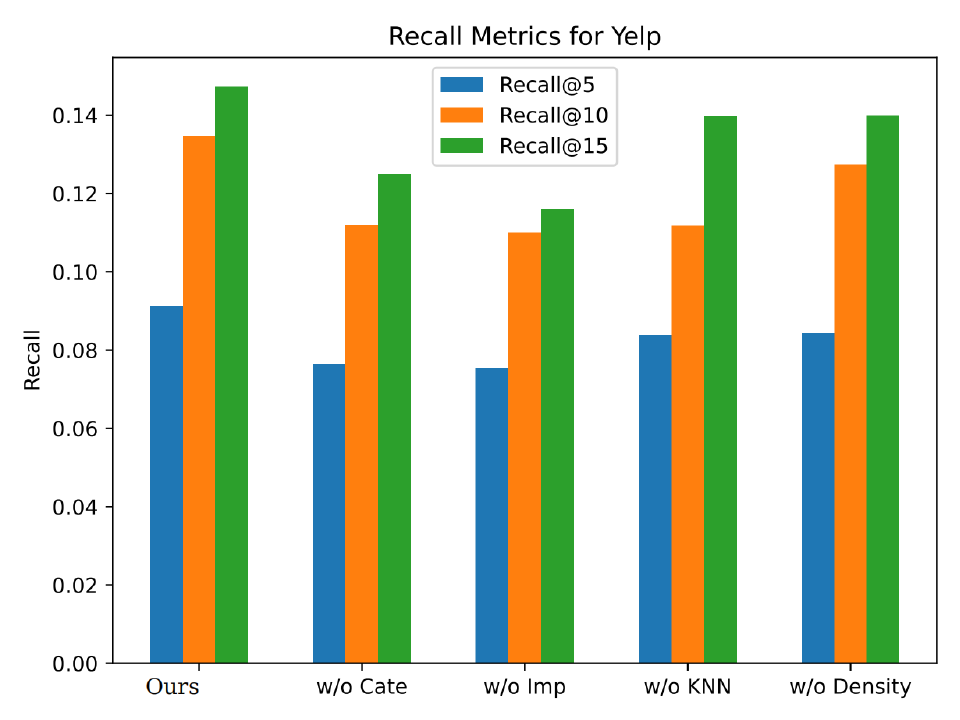}
%\caption{fig2}
\end{minipage}}
\subfigure[GOT by foursquare]
{\begin{minipage}[t]{0.48\linewidth}
\centering
\includegraphics[scale=0.26]{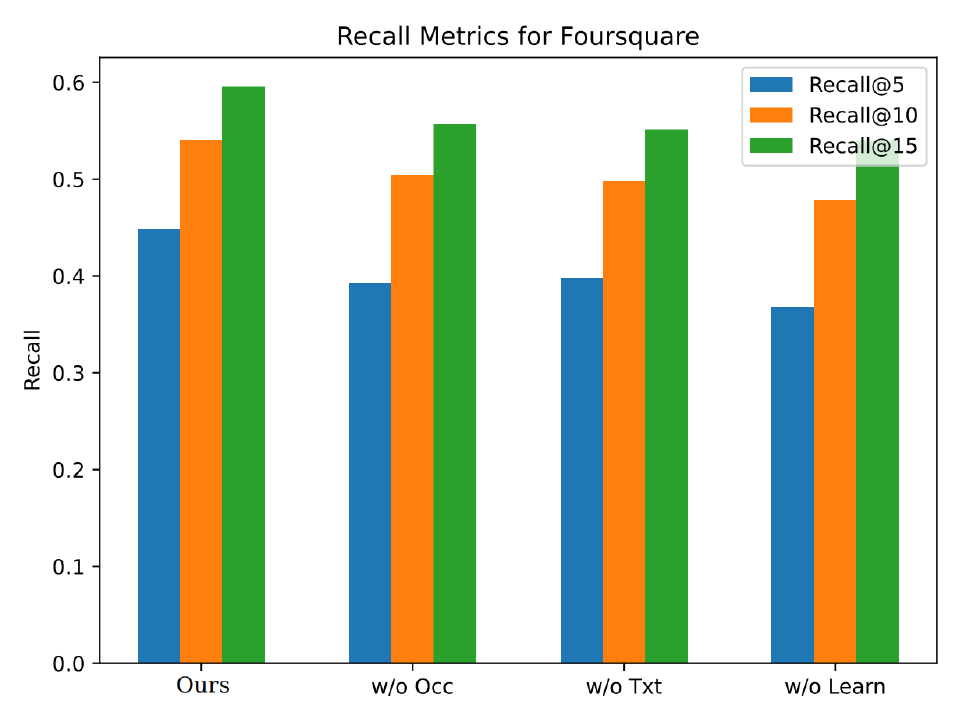}
%\caption{fig2}
\end{minipage}}
\subfigure[GOT by Yelp]
{\begin{minipage}[t]{0.48\linewidth}
\centering
\includegraphics[scale=0.26]{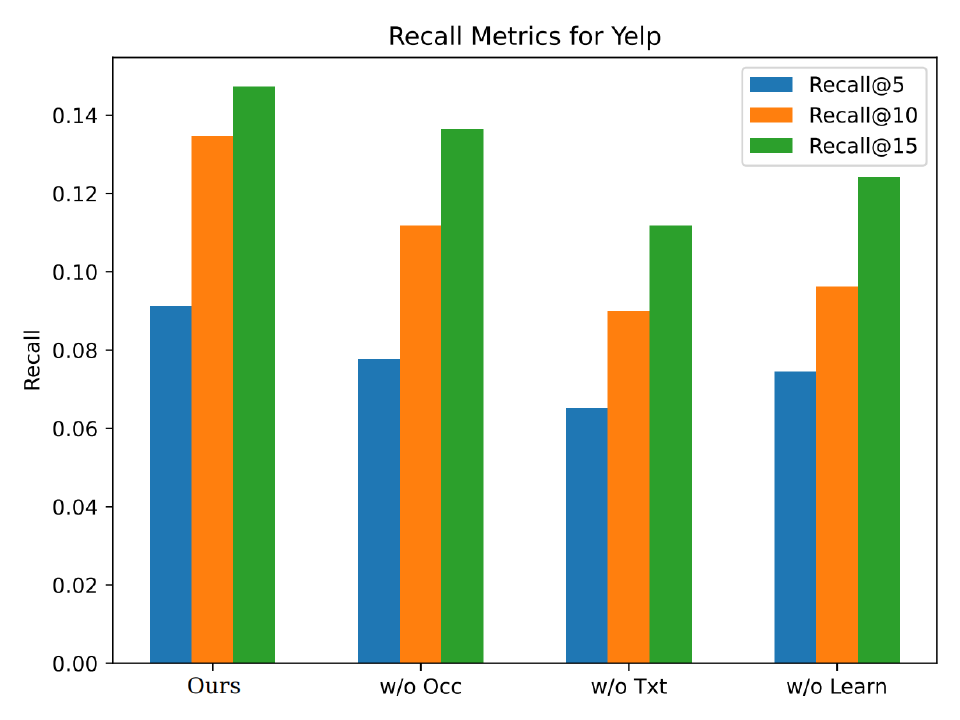}
%\caption{fig2}
\end{minipage}}

\caption{Ablation study on mixed sampling and GOT representation.}
\label{fig:Ablation study}
\vspace{-2em}
\end{figure}

Figure~\ref{fig:Ablation study} highlights the indispensability of each component in the mixed sampling and GOT frameworks. Removing any element reduces recall at @5, @10, and @15. In Figure~\ref{fig:Ablation study} (a), excluding density-based sampling or KNN lowers Recall@5 by 7.6\% and 7.7\%, while removing category-aware or importance sampling reduces Recall@10 by 3.3\% and 5.0\%. Figure~\ref{fig:Ablation study} (b) shows average recall drops of 19.0\% and 10.1\% without category-aware or importance sampling. 

Figure~\ref{fig:Ablation study} (c) reveals that removing co-occurrence, text representation, or all components reduces Recall@5 by 12.6\%, 11.2\%, and 17.9\%, respectively, while freezing embeddings causes the largest drop. In Figure~\ref{fig:Ablation study} (d), omitting text encoding cuts average recall by 28.6\%. These results highlight the critical role of multi-contextual and text information in POI embeddings.

\subsubsection{Adaptive module and Attention mechanism}
\label{sec:Ablation_att}

We show the ablation study on adaptive and attention module by Table~\ref{table:ablation_nextpoi}. Specifically, we remove the component with the adaptive representation aggregator by w/o ARA, Geographical component by w/o G, the Co-occurrence component by w/o O, and the Textual component by w/o T. We also replace GOTAtt with Vanilla Attention by GOTAtt.

\begin{table}[t]
\vspace{-1em}
\caption{Ablation study on Adaptive Representation Aggregator (ARA) and attention mechanism.}
\centering
%\scriptsize % Smaller font size for one-column layout
\setlength{\tabcolsep}{1pt} % Reduce column spacing
\renewcommand{\arraystretch}{0.9} % Adjust row spacing
\resizebox{\linewidth}{!}{
\begin{tabular}{ll!{\vrule width 0.2pt}lll|lll}
\hline
\multicolumn{2}{c!{\vrule width 0.2pt}}{Dataset} & \multicolumn{3}{c|}{NY-Next POI rec} & \multicolumn{3}{c}{LA-Next POI rec} \\ \hline
\multicolumn{1}{c}{DM} & \multicolumn{1}{c!{\vrule width 0.2pt}}{PM} & \multicolumn{1}{c}{Rec@5} & \multicolumn{1}{c}{Rec@10} & \multicolumn{1}{c|}{Rec@15} & \multicolumn{1}{c}{Rec@5} & \multicolumn{1}{c}{Rec@10} & \multicolumn{1}{c}{Rec@15} \\ \hline
 & GOTAtt & 0.278 & 0.389 & 0.448 & 0.062 & 0.103 & 0.127 \\
 & w/o ARA & 0.275& 0.384 & 0.441 & 0.06 & 0.098 & 0.123\\
 & w/o G & 0.284 & 0.401 & 0.463 & 0.068 & 0.111 & 0.134 \\ 
 & w/o O & 0.295 & 0.408 & 0.468 & 0.073 & 0.118 & 0.141 \\ 
 & w/o T & 0.286 & 0.403 & 0.459 & 0.069 & 0.115 & 0.136 \\ 
\multirow{-5}{*}{FPMC} & \textbf{GOT} & \textbf{0.312} & \textbf{0.423} & \textbf{0.486} & \textbf{0.086} & \textbf{0.125} & \textbf{0.156} \\ \hline
 & GOTAtt & 0.396 & 0.492 & 0.526 & 0.061 & 0.095 & 0.103 \\ 
& w/o ARA & 0.392& 0.487 & 0.519 & 0.058 & 0.095 & 0.102\\ 
 & w/o G & 0.411 & 0.504 & 0.539 & 0.071 & 0.103 & 0.114 \\ 
 & w/o O & 0.415 & 0.511 & 0.546 & 0.073 & 0.107 & 0.116 \\ 
 & w/o T & 0.405 & 0.496 & 0.538 & 0.067 & 0.108 & 0.115 \\ 
\multirow{-5}{*}{LSTPM} & \textbf{GOT} & \textbf{0.426} & \textbf{0.513} & \textbf{0.557} & \textbf{0.078} & \textbf{0.112} & \textbf{0.121} \\ \hline
\end{tabular}
}
\label{table:ablation_nextpoi}
\vspace{-1em}
\end{table}
The results first demonstrate the critical role of the adaptive component. Removing ARA (w/o ARA) consistently leads to performance degradation across both datasets and baselines, underscoring its importance in adaptively aggregating subgraphs. In addition, the effectiveness of the proposed GOT attention mechanism is clearly validated by comparisons with its ablated variants (GOTAtt, w/o G, w/o O, and w/o T). GOT consistently outperforms these alternatives in terms of recall. Notably, under the FPMC model on the NY dataset, GOT achieves a 12.2\% improvement in Recall@5 (from 0.278 to 0.312) and an 8.5\% gain in Recall@15 (from 0.448 to 0.486), further confirming the value of GOT attention mechanism.
\subsection{Case Studies}
\label{sec:Case}

\begin{figure}[h]
\vspace{-1em}
\centering
\subfigure[POIs]
{
\begin{minipage}[t]{0.4\linewidth} % Adjusted width for even distribution
\centering
\includegraphics[scale=0.18]{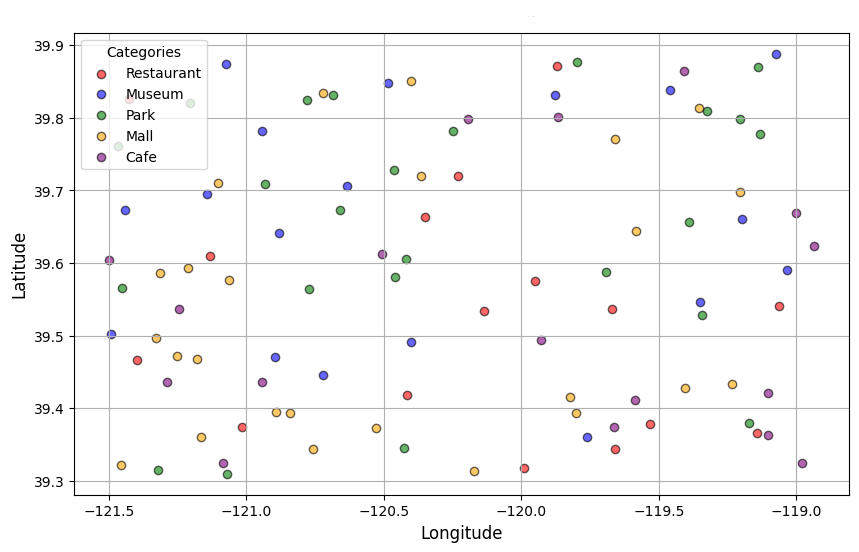} % Slightly adjusted scale
\end{minipage}}
\hfill % Ensures even spacing
\subfigure[ BNS~\cite{bns}]
{\begin{minipage}[t]{0.4\linewidth}
\centering
\includegraphics[scale=0.18]{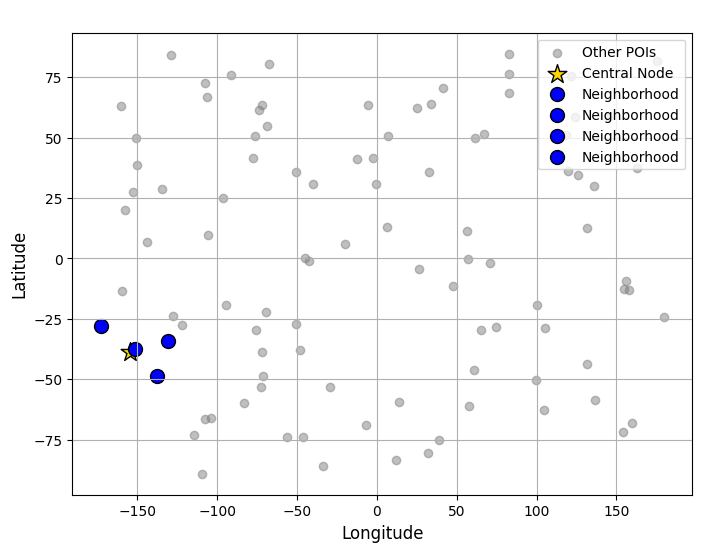}
\end{minipage}}
\vspace{0.2cm} % Adds vertical spacing between rows
\subfigure[AdaptGOT]
{\begin{minipage}[t]{0.48\linewidth}
\centering
\includegraphics[scale=0.18]{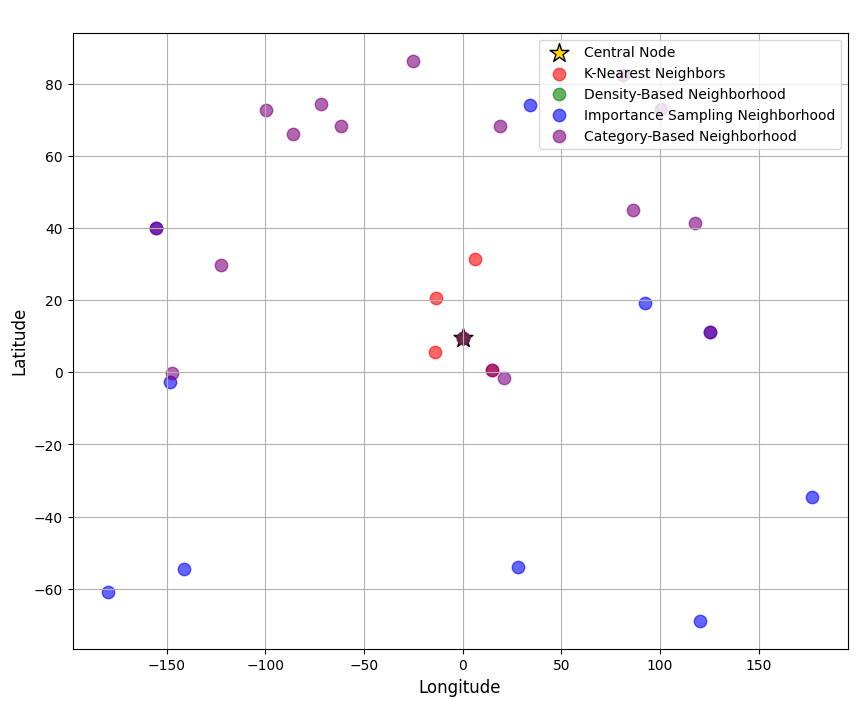}
\end{minipage}}
\hfill
\subfigure[Visualization of embedding space]
{\begin{minipage}[t]{0.48\linewidth}
\centering
\includegraphics[scale=0.18]{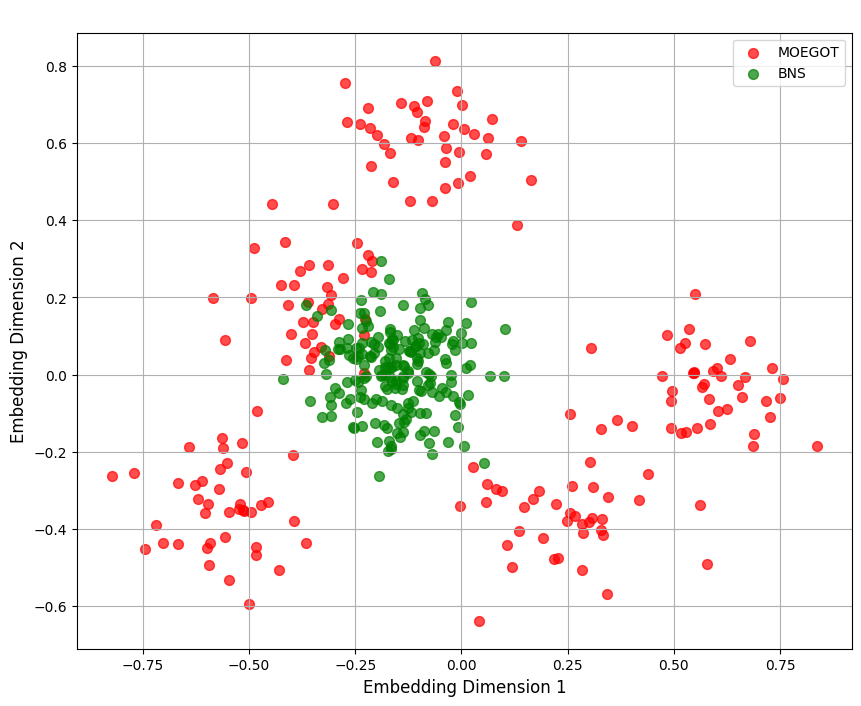}
\end{minipage}}
\caption{Case study on the Yelp dataset. Each subfigure represents key insights about POI selection, neighborhood generation, and embedding space exploration.}
\label{fig:Casestudy}
\vspace{-1em}
\end{figure}

This case study addresses two questions: 1) Can AdaptGOT learn related subgraphs through mixed sampling? 2) Can AdaptGOT identify GOT contexts? Using 100 randomly selected POIs from the Yelp dataset as Figure~\ref{fig:Casestudy} (a) shows, we analysed subgraphs and embeddings. Figure~\ref{fig:Casestudy} (b) shows BNS-generated subgraphs~\cite{bns}, which focus on proximal neighbors but ignore broader contextual factors. In contrast, Figure~\ref{fig:Casestudy} (c) highlights AdaptGOT’s ability to integrate category, density, and proximity, creating robust subgraphs. Figure~\ref{fig:Casestudy} (d) compares embedding spaces, where AdaptGOT produces globally uniform embeddings that absorb diverse contexts, while BNS suffers from over-smoothing and clustered representations. Overall, AdaptGOT effectively captures contextual information, outperforming BNS in subgraph and embedding quality.

\subsection{Sensitivity Studies}
\label{sec:Sensitivity}

In this section, we study the impact of parameters for \textsf{AdaptGOT} on the next POI recommendation task and use recall, F1 score, Mean Reciprocal Rank (MRR), and Normalized Discounted Cumulative Gain (NDCG) as performance metrics.

\subsubsection{Number of subgraphs}
We evaluated subgraphs ranging from 2 to 6, using random combinations of sampling methods for two or three subgraphs\footnote{We claim that when the number of subgraphs above 4, additional subgraph structures are constructed by different parameters (e.g. different $k$ in KNN) to prevent the over-reliance on redundant topologies.} . Performance peaked with four subgraphs, offering the best balance of effectiveness and computational efficiency in Table~\ref{tab:sensi_subgraph}. While additional subgraphs slightly improved performance, the gains were marginal and incurred higher computational costs. Therefore, we chose four subgraphs as optimal for both datasets.

\begin{table}[ht]
\vspace{-1em}
\centering
\caption{Sensitivity to the number of subgraphs in FPMC~\cite{FPMC} for next POI recommendation.}
\scriptsize
\setlength{\tabcolsep}{3pt} % Adjust column spacing
\renewcommand{\arraystretch}{0.9} % Adjust row spacing
\resizebox{\linewidth}{!}{
\begin{tabular}{cc!{\vrule width 0.3pt}ccc|ccc}
\hline
\multicolumn{2}{c!{\vrule width 0.3pt}}{Dataset} & Rec@5 & Rec@10 & Rec@15 & Rec@5 & Rec@10 & Rec@15 \\ \hline
\multicolumn{1}{c}{DM}  & \multicolumn{1}{c!{\vrule width 0.3pt}}{Num} 
& \multicolumn{3}{c|}{NY-Next POI rec} & \multicolumn{3}{c}{LA-Next POI rec} \\ \hline
 & 2 & 0.298 & 0.411 & 0.472 & 0.073 & 0.107 & 0.142 \\
 & 3 & 0.307 & 0.418 & 0.479 & 0.081 & 0.118 & 0.149 \\
 & 4 & 0.312 & 0.423 & \textbf{0.486} & \textbf{0.086} & \textbf{0.125} & \textbf{0.156} \\
 & 5 & \textbf{0.313} & \textbf{0.426} & 0.483 & 0.085 & 0.123 & 0.154 \\
\multirow{-5}{*}{FPMC} & 6 & 0.312 & 0.423 & 0.483 & \textbf{0.087} & 0.124 & 0.155 \\
\hline
\end{tabular}
}
\label{tab:sensi_subgraph}
\vspace{-1em}
\end{table}

\subsubsection{Number of Neighbours Connections}

\begin{figure}[h]
    \centering
    \includegraphics[width=0.9\linewidth]{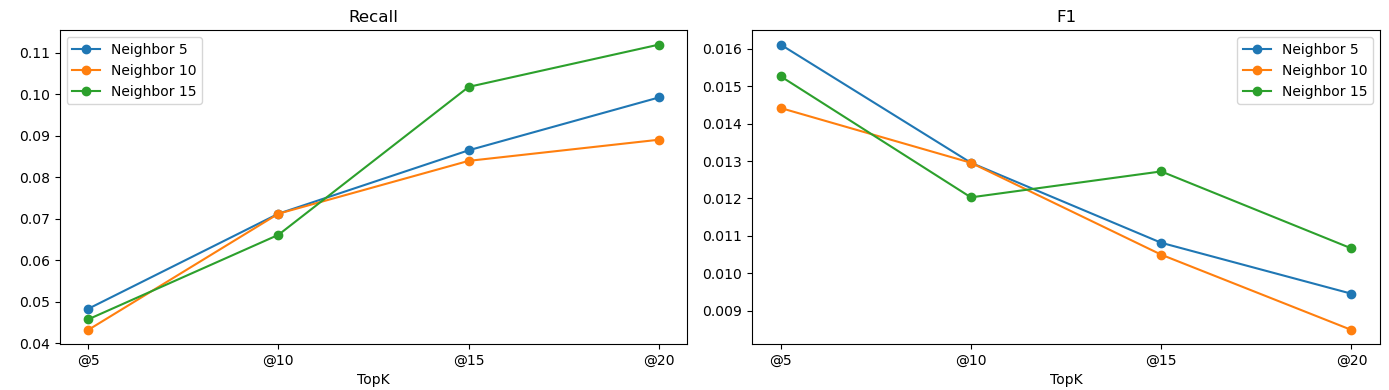}
    \caption{Sensitivity studies on different number of neighbours connection.}
    \label{fig:num-neighbors}
\end{figure}

To examine the impact of varying the number of neighbour connections during the graph construction phase in pre-training on downstream task performance, we employ the HST-LSTM for the next POI recommendation task on Yelp LA with the number of neighbours of 5, 10 and 15. As figure \ref{fig:num-neighbors} shows, 15 neighborhoods outperform others in the majority cases. 
% This highlights the efficacy of increased neighbor connections in ranking relevant POIs within subgraph instances across the recommendation list. However,  employing more neighbors can inevitably incur additional computational costs, contradicting the original intent of our subgraph design.

% \begin{table*}[h]
% \centering
% \caption{Result on the HST-LSTM model for Next POI recommendation on Yelp Louisana with a different number of neighbours connection.}
% \resizebox{\textwidth}{!}{%
% \begin{tabular}{l|cccc|cccc|cccc}
% \hline
%  & \multicolumn{4}{c|}{Neighbor 5} & \multicolumn{4}{c|}{Neighbor 10} & \multicolumn{4}{c}{Neighbor 15} \\
% Metric & @5 & @10 & @15 & @20 & @5 & @10 & @15 & @20 & @5 & @10 & @15 & @20 \\ 
% \hline
% Recall & 0.0483 & 0.0712 & 0.0865 & 0.0992 & 0.0432 & 0.0712 & 0.0839 & 0.0890 & 0.0458 & 0.0662 & 0.1018 & 0.1119 \\
% F1     & 0.0161 & 0.0129 & 0.0108 & 0.0095 & 0.0144 & 0.0129 & 0.0105 & 0.0085 & 0.0153 & 0.0120 & 0.0127 & 0.0107 \\
% MRR    & 0.0255 & 0.0281 & 0.0293 & 0.0301 & 0.0264 & 0.0301 & 0.0310 & 0.0313 & 0.0292 & 0.0316 & 0.0344 & 0.0350 \\
% NDCG   & 0.0311 & 0.0381 & 0.0422 & 0.0452 & 0.0307 & 0.0398 & 0.0430 & 0.0442 & 0.0333 & 0.0396 & 0.0490 & 0.0514 \\
% \hline
% \end{tabular}
% }
% \label{table:neighbors-metrics}
% \end{table*}

% When increasing from 5 to 10 numbers of neighbours, the recall only improves when K is 5, but not across other K values. However, for the MRR, the 10 neighbours setting is outperformed across all K values. It suggests 10 neighbours is better at placing a relevant POI higher in the recommendation list than  5 neighbours, regardless of the list's length. 15 neighbours also outperform 10 neighbours therefore more numbers of neighbours connected might be preferred when the priority is to ensure that the top few recommendations (especially the very first one) are highly relevant, given its superior MRR. NDCG scores increase with the number of neighbours, with the 15 neighbours setup outperforming both 5 and 10 across all K values. This improvement highlights the effectiveness of more neighbour connections in ranking relevant POIs higher across the recommendation list.
% A crucial consideration is the significant increase in computational costs associated with higher numbers of neighbour connections. This is because the model must aggregate and process a more extensive context for each POI, leading a high memory cost for the pre-training phase. A higher number of neighbour connections makes the graph denser, which directly increases the memory required to store the graph. This includes the adjacency matrix, which grows quadratically with the number of edges, and the edge feature vectors.
\subsubsection{Number of heads in GOT Attention}

\begin{figure}[h]
    \centering
    \includegraphics[width=0.9\linewidth]{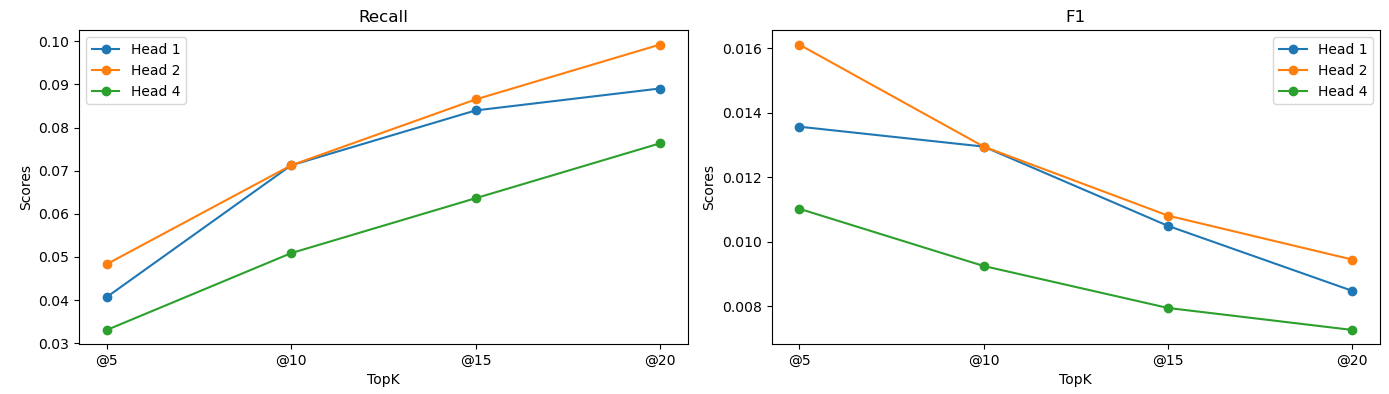}
    \caption{Sensitivity study on different heads numbers.}
    \label{fig:multi-head}
\end{figure}

In the GOT Attention mechanism, we assess the impact of varying the number of heads in multi-head attention by selecting 1, 2, or 4 heads. We employ the HST-LSTM model \cite{hstlstm} as the downstream model for the next POI recommendation task. As depicted in Figure \ref{fig:multi-head}, the configuration with 2 heads demonstrates superior performance.

% \begin{table*}[h]
% \centering
% \caption{RQ4: Sensitivity studies - Number of heads using the HSTLSTM model for next POI recommendation}
% % using the HSTLSTM model for next poi rec
% % number of head when neighbor is 5
% %yelp_la
% \resizebox{\textwidth}{!}{%
% \begin{tabular}{l|cccc}
% \hline
% Variants & Recall@5 & Recall@10 & Recall@15 & Recall@20 \\ \hline
% head-1 & 0.040712 & 0.0712468 & 0.083969 & 0.08905852 \\
% head-2 & 0.04834605 & 0.071246819 & 0.086513994 & 0.09923664122 \\
% head-4 & 0.03307888040 & 0.05089058 & 0.06361323 & 0.076335877862 \\
% \hline
% \end{tabular}
% }
% \label{table:heads}
% \end{table*}

\subsection{Zero-shot Cross-city Generalisation}
\label{sec:othertask}

% We want to explore further the effectiveness of MOEGOT on other tasks, such as transferability. 
% We evaluate the transferability performance of \textsf{MOEGOT} in cross-city tasks. Take the Foursquare dataset as an example. We pre-trained the model using Foursquare Tokyo as the source city, then froze the model and subsequently utilized data from New York (NY) to generate the POI embeddings. Likewise, we pre-trained the model using data from Yelp NV and test on Yelp LA. 

% Table \ref{table:random-pretrain-comparison} shows the cross-city transferability performance of pre-trained embeddings on the next POI recommendation task. Overall, \textsf{MOEGOT} exhibited a notable increase in an average recall by 19.7\%, and it demonstrates significant enhancements over both Random and SpaBERT embeddings on the Yelp LA dataset, showcasing a noteworthy 14.9\% and 20.1\% improvement over Random in Recall@5 and Recall@10, respectively. Moreover, there is a consistent decline in performance for SpaBERT compared to Random across nearly all models on both datasets. This means SpaBERT cannot adapt to new domains with unseen data. These results underscore \textsf{MOEGOT}'s superior ability to learn a unified representation through POI contextual information from multiple context graphs, thereby enabling improved transferability to cross-city settings.
% However, SpaBERT fails to perform well on the Yelp dataset on both FPMC~\cite{FPMC} and LSTPM~\cite{LSTPM}, showing the instability of capturing contextual information on cross-city tasks.

We evaluated the cross-city transferability of \textsf{AdaptGOT} using Foursquare and Yelp datasets. 
% For Foursquare, the model was pre-trained on Tokyo and tested on New York, while for Yelp, it was pre-trained on NV and tested on LA. 
Table~\ref{table:random-pretrain-comparison} highlights \textsf{AdaptGOT}'s superior performance, achieving an average recall increase of 19.7\% and outperforming Random and SpaBERT embeddings on the Yelp LA dataset, with 14.9\% and 20.1\% improvements in Recall@5 and Recall@10, respectively.

% Notably, SpaBERT underperformed compared to Random across most models and datasets, indicating its inability to adapt to unseen domains. These results confirm \textsf{AdaptGOT}'s ability to leverage multi-contextual information for unified representations, enabling robust transferability in cross-city tasks.

\begin{table}[!h]
\centering
\caption{Cross-City Transferability Performance on Next POI Recommendation.}
\label{table:random-pretrain-comparison}
\footnotesize
\renewcommand{\arraystretch}{1.0}
\setlength{\tabcolsep}{0.9mm}
\resizebox{\linewidth}{!}{
\begin{tabular}{ll|ccc|ccc}
\hline
\multicolumn{2}{c|}{Dataset} & \multicolumn{3}{c|}{Foursquare} & \multicolumn{3}{c}{Yelp} \\ \hline
DM   & PM   & Rec@5 & Rec@10 & Rec@15 & Rec@5 & Rec@10 & Rec@15 \\ \hline
FPMC  & Rand & 0.283 & 0.394 & 0.461 & 0.072 & 0.087 & 0.101 \\
FPMC  & FC   & 0.275 & 0.388 & 0.437 & 0.059 & 0.074 & 0.087 \\
FPMC  & CTLE & 0.286 & 0.401 & 0.464 & 0.075 & 0.096 & 0.115 \\
FPMC  & Spa  & 0.242 & 0.340 & 0.397 & 0.034 & 0.058 & 0.073 \\
FPMC  & GOT  & \textbf{0.293} & \textbf{0.414} & \textbf{0.488} & \textbf{0.082} & \textbf{0.111} & \textbf{0.130} \\ \hline
LSTPM & Rand & 0.399 & 0.485 & 0.529 & 0.064 & 0.079 & 0.089 \\
LSTPM & FC   & 0.394 & 0.490 & 0.533 & 0.069 & 0.084 & 0.092 \\
LSTPM & CTLE & 0.395 & 0.471 & 0.504 & 0.059 & 0.064 & 0.078 \\
LSTPM & Spa  & 0.399 & 0.479 & 0.519 & 0.045 & 0.069 & 0.084 \\
LSTPM & GOT  & \textbf{0.408} & \textbf{0.497} & \textbf{0.548} & \textbf{0.075} & \textbf{0.089} & \textbf{0.097} \\ \hline
\end{tabular}
}
\end{table}

\section{Conclusion}

We presented \textsf{AdaptGOT}, a novel framework addressing key challenges in POI representation learning, such as multi-context sampling strategies, insufficient exploration of multiple POI contexts, and limited adaptability to diverse tasks. By integrating an adaptive learning component with Geographical-Co-Occurrence-Text (GOT) representations, \textsf{AdaptGOT} enhances the expressive power and versatility of POI embeddings. 
%The framework comprises three core components: a contextual neighborhood generation scheme leveraging mixed sampling methods to capture diverse contextual neighborhoods; a GOT representation module with an attention mechanism for deriving customized representations and modeling intricate POI interrelations; and an MOE encoder-decoder architecture that ensures topological consistency and improves contextual representation by minimizing Jensen-Shannon divergence. 
Extensive experiments demonstrated the superior performance of \textsf{AdaptGOT} compared to state-of-the-art POI embedding methods. 
% Future directions include extending \textsf{MOEGOT} to incorporate temporal dynamics and exploring its scalability to larger datasets and new application domains. 

\vspace{8pt} % Slightly reduces the space, adjust as needed
% \bibliographystyle{ACM-Reference-Format}
\vspace{8pt} % Slightly reduces the space, adjust as needed


\end{document}